\documentclass[sn-mathphys,Numbered]{sn-jnl}

\usepackage{longtable}
\usepackage{graphicx}%
\usepackage{multirow}%
\usepackage{amsmath,amssymb,amsfonts}%
\usepackage{amsthm}%
\usepackage{mathrsfs}%
\usepackage[title]{appendix}%
\usepackage[table,xcdraw]{xcolor}
\usepackage{textcomp}%
\usepackage{manyfoot}%
\usepackage{booktabs}%
\usepackage{algorithm}%
\usepackage{algorithmicx}%
\usepackage{algpseudocode}%
\usepackage{listings}%
\usepackage{comment}
\usepackage{verbatim}

\usepackage{hyperref}
\usepackage{threeparttable}
\usepackage{rotating}
\usepackage{comment}

\usepackage{pdflscape}
\usepackage{longtable}
\usepackage{rotating}
\def\Rot#1{ \multicolumn{1}{l|}{\rlap{\rotatebox{90}{#1}~}}}
\def\Rott#1{ \multicolumn{1}{l}{\rlap{\rotatebox{45}{#1}~}}}

\usepackage[inline]{enumitem}

\usepackage{rotating}
\def\Rot#1{ \multicolumn{1}{l|}{\rlap{\rotatebox{90}{#1}~}}}

\usepackage{array}
\newcolumntype{L}[1]{>{\raggedright\let\newline\\\arraybackslash\hspace{0pt}}m{#1}}
\newcolumntype{C}[1]{>{\centering\let\newline\\\arraybackslash\hspace{0pt}}m{#1}}
\newcolumntype{R}[1]{>{\raggedleft\let\newline\\\arraybackslash\hspace{0pt}}m{#1}}
\usepackage{booktabs, makecell, threeparttable, tabularx, boldline}

\usepackage{caption}
\usepackage{subcaption}

\newcommand\changed[1]{{\color{black}{#1}}}
\newcommand\changedtwo[1]{{\color{black}{#1}}}

\theoremstyle{thmstyleone}%
\theoremstyle{thmstyletwo}%

\theoremstyle{thmstylethree}%

\begin{document}

\title{Artificial Intelligence for Literature Reviews: Opportunities and Challenges}
%

\author*[1]{\fnm{Francisco} \sur{Bolaños}}\email{francisco.bolanos-burgos@open.ac.uk}

\author[1]{\fnm{Angelo} \sur{Salatino}}\email{angelo.salatino@open.ac.uk}

\author[1,2]{\fnm{Francesco} \sur{Osborne}}\email{francesco.osborne@open.ac.uk}

\author[1]{\fnm{Enrico} \sur{Motta}}\email{enrico.motta@open.ac.uk}

\affil[1]{\orgdiv{Knowledge Media Institute}, \orgname{The Open University}, \orgaddress{\street{Walton Hall}, \city{Milton Keynes}, \postcode{MK7 6AA}, \country{UK}}}

\affil[2]{\orgdiv{Department of Business and Law}, \orgname{University of Milano Bicocca}, \orgaddress{\street{Piazza dell'Ateneo Nuovo, 1}, \city{Milan}, \postcode{20126}, \country{IT}}}


\abstract{
This paper presents a comprehensive review of the use of Artificial Intelligence (AI) in Systematic Literature Reviews (SLRs). A SLR is a rigorous and organised methodology that assesses and integrates prior research on a given topic. Numerous tools have been developed to assist and partially automate the SLR process. The increasing role of AI in this field shows great potential in providing more effective support for researchers, moving towards the semi-automatic creation of literature reviews. Our study focuses on how AI techniques are applied in the semi-automation of SLRs, specifically in the screening and extraction phases. We examine 21 leading SLR tools using a framework that combines 23 traditional features with 11 AI features. We also analyse 11 recent tools that leverage large language models for searching the literature and assisting academic writing. Finally, the paper discusses current trends in the field, outlines key research challenges, and suggests directions for future research. We highlight three primary research challenges: integrating advanced AI solutions, such as large language models and knowledge graphs, improving usability, and developing a standardised evaluation framework. 
We also propose best practices to ensure more robust evaluations in terms of performance, usability, and transparency. Overall, this review offers a detailed overview of AI-enhanced SLR tools for researchers and practitioners, providing a foundation for the development of next-generation AI solutions in this field.
}




\maketitle

\section{Introduction}\label{introduction}


A Systematic Literature Review (SLR) is a rigorous and organised methodology that assesses and integrates previous research on a specific topic. Its main goal is to meticulously identify and appraise all the relevant literature related to a specific research question, adhering to strict protocols to minimise biases~\cite{higgins2011cochrane,moher2009preferred}. 
This methodology originally emerged within the realm of Evidence-Based Medicine~\cite{sackett1996evidence}, and it was subsequently adapted and employed in diverse research disciplines including social sciences~\cite{petticrew2008systematic}, engineering and technology~\cite{keele2007guidelines}, education~\cite{gough2017introduction}, environmental sciences~\cite{pullin2006guidelines}, and business and management~\cite{tranfield2003towards}.


SLRs are recognised for being time-consuming and resource-intensive. 
This is due to several factors, including the lengthy process that can extend beyond a year~\cite{borah2017analysis}, the necessity of assembling a team of domain experts~\cite{shojania2007quickly}, significant financial implications from database subscriptions, specialised software, and personnel remuneration~\cite{shemilt2016use}, the growing number of publications~\cite{bornmann2015}, and the periodic need for updates to maintain relevance~\cite{moher2007systematic}. 

Over the past decades, numerous tools have been developed to support and even partially automate SLRs, aiming to address these challenges. 
Many of these tools have adopted Artificial Intelligence (AI) solutions~\cite{van2022automatic,kebede2023depth}, particularly for the screening and data extraction phases. 
The incorporation of AI into SLR tools has been further propelled by the emergence of more sophisticated AI techniques in Natural Language Processing (NLP), such as Large Language Models (LLMs), which have the potential to revolutionise these systems~\cite{robinson2023bio}. 
While a significant body of research has examined SLR tools~\cite{carver2013identifying,feng2017text,napoleao2021automated,wagner2022artificial,khalil2022tools,cierco2022machine,TulkJesso2022InclusionOC,robledo2023vista,Ng2023AttitudesAP}, relatively few studies have explored the role of AI in this domain~\cite{cowie2022web,de2023artificial,burgard2023reducing,robledo2023vista}. Furthermore, these studies focused on a limited selection of AI features, as we will discuss in Section~\ref{sec:metareview}.



This survey aims to address the existing gap by rigorously examining the application of AI techniques in the semi-automation of SLRs, within the two main stages of application, namely \textit{screening} and \textit{extraction}.
For this purpose, we first conducted an analysis of eight prior surveys and identified the most prominent features examined in the literature. Next, we defined a framework of analysis that integrates 23 general features and 11 features pertinent to AI-based functionalities. We then selected 21 prominent SLR tools and subjected them to rigorous analysis using the resulting framework. \changedtwo{We extensively discuss current trends, key research challenges, and directions for future research. We specifically focus on three major research challenges: 1) integrating advanced AI solutions, such as large language models and knowledge graphs, 2) enhancing usability, and 3) developing a standardised evaluation framework. We also propose a set of best practices to ensure more robust evaluations regarding performance, usability, and transparency. Finally, we performed an additional analysis on 11 recent tools that utilise the capabilities of  LLMs (predominantly ChatGPT via the OpenAI API) for searching the literature and aiding academic writing.} Although these tools do not cater directly to SLRs, there is potential for their features to be integrated into future SLR tools. 
In conclusion, this survey seeks to offer scholars a thorough insight into the application of Artificial Intelligence in this field, while also highlighting potential avenues for future research.


The remainder of this paper is structured as follows. \changed{Section~\ref{sec:Background} includes a description of the SLR stages and their relationship with AI. 
Section~\ref{sec:toolselection} outlines the methodology we employed to identify the SLR tools discussed in the survey. 
Section~\ref{sec:metareview} provides a meta-review of previous surveys about SLR tools that analysed AI features.
Section~\ref{oursurvey} provides an in-depth examination of the 21 tools. 
Section~\ref{sec:challenges} discusses the key research challenges and proposes some best practices for the evaluation of AI-enhanced SLR tools. 
Section~\ref{sec:other_ai_tools} analyses the latest generation of LLM-based systems designed to assist researchers. 
Finally, Section~\ref{conclusion} concludes the paper by summarising the contributions and the main findings. }

\section{Background}\label{sec:Background}





In this section, we examine the various stages of a SLR and the extent of support they receive from AI in the current generation of tools. Here, the term `AI' specifically denotes weak or narrow AI, which includes systems designed and trained for specific tasks like classification, clustering, or named-entity recognition~\cite{iansiti2020competing}. In the context of SLR, these methodologies are predominantly utilised to semi-automate tasks like screening and data extraction~\cite{cowie2022web,burgard2023reducing}.

The SLR methodology consists of six distinct stages~\cite{higgins2011cochrane,keele2007guidelines}: 
\begin{enumerate*}[label=\roman*)]
    \item Planning,
    \item Search,
    \item Screening,
    \item Data Extraction and Synthesis,
    \item Quality Assessment, and
    \item Reporting.
\end{enumerate*}
Each stage plays a pivotal role in ensuring the comprehensiveness and rigour of the review process.

The \textit{planning} phase is foundational to the entire review process, as it involves formulating a set of precise and specific research questions that the SLR seeks to address~\cite{o2008defining}. A detailed protocol is also developed during this stage, outlining the appropriate methodologies that will be adopted to carry out the review~\cite{fontaine2022designing}. This protocol ensures consistency, reduces bias, and enhances the transparency and reproducibility of the review.

The \textit{search} phase aims to identify relevant papers using search strategies, snowballing, or a hybrid approach. Search strategies are typically implemented by creating a query based on a combination of terms using boolean operators~\cite{glanville2019development,Team2007InformationRG}. This query is then executed on designated search engines.
In snowballing, the researcher examines the references and citations of an initial group of papers (also known as seed papers) to identify additional articles. This process is iteratively repeated until no new relevant scholarly documents are found~\cite{webster2002analyzing,wohlin2014guidelines}. The hybrid approach is the combination of search strategy and snowballing~\cite{mourao2017investigating,wohlin2022successful}. 
Traditionally, the search phase had not been significantly supported by artificial intelligence techniques~\cite{adam2022semi}. Nevertheless, there are some emerging tools, which we will examine in Section \ref{sec:other_ai_tools}, that have begun to incorporate LLMs in academic search engines, often within a Retrieval-Augmented Generation (RAG) framework~\cite{lewis2020retrieval}. This innovative approach allows for the formulation of precise questions and complex queries in natural language, surpassing the capabilities of traditional keyword-based searches.

The \textit{screening} phase uses a set of inclusion and exclusion criteria to further filter the paper obtained from the search stage.
It typically consists of two stages: i) title and abstract screening and ii) full-text screening. In the first step,  the reviewers screen the relevant papers according only to the title and abstract~\cite{moher2009preferred}. The second step entails a detailed evaluation of the content of each paper, a task that demands significantly more effort but leads to a more thorough assessment. It is also customary to document the rationale for excluding any given paper during this process. 
The predominant application of AI in SLR regards this phase. It usually involves employing machine learning classifiers, which are trained on an initial set of user-selected papers and then used to identify additional relevant articles~\cite{miwa2014reducing}. This process frequently involves iteration, where the user refines the automatic classifications or selects new papers, followed by retraining the classifier to better identify further pertinent literature.

In the \textit{data extraction and synthesis} phase, all the pertinent information is systematically extracted from the selected studies. The techniques for data extraction vary greatly depending on the research field and the objective of the researcher. For example, in the biomedical field, protocols like PECODR~\cite{dawes2007identification} (Patient-Population-Problem, Exposure-Intervention, Comparison, Outcome, Duration, and Results) and PIBOSO~\cite{kim2011automatic} (Population, Intervention, Background, Outcome, Study Design, and Other) are used to identify key elements from clinical studies, while the STARD checklist~\cite{bossuyt2003towards} supports readers in assessing the risk of bias and evaluating the relevance of the results.
Following the extraction, the data is aggregated and summarised~\cite{garousi2017experience,munn2014jbi}. Depending on the nature and heterogeneity of the data, the resulting synthesis might be qualitative or quantitative. 
This phase is also occasionally supported by AI solutions. Commonly, the relevant tools employ classifiers to identify articles possessing specific characteristics~\cite{marshall2018machine} or implement named-entity recognition for extracting specific entities or concepts~\cite{kiritchenko2010exact} (e.g., RCT entities~\cite{moher2001consort},  entities pertaining environmental health studies~\cite{walker2022evaluation}).

The \textit{quality assessment} phase evaluates the rigour and validity of the selected studies~\cite{higgins2008assessing,wells2000newcastle,von2007strengthening,effective1998quality}. This analysis provides evidence of the overall strength and the level of trustworthiness presented in the review~\cite{zhou2015quality,chen2022grading}. 

Finally, the \textit{reporting} phase involves presenting the findings in a structured and coherent manner within a research paper. This presentation typically follows an established format comprising sections like introduction, methods, results, and discussion, but this may differ depending on the journal in which the manuscript will be published~\cite{page2021prisma,stroup2000meta}. 
Historically, this stage did not benefit from the use of artificial intelligence techniques~\cite{li2022automatic,justitia2022automatic}. However, as we will discuss in Section \ref{sec:other_ai_tools}, recent advancements have led to the development of tools based on LLMs designed to support academic writing, which can be particularly useful in this phase. These tools typically enable users to draft an initial outline of the desired document and iteratively refine it.

\section{Methodology}\label{sec:toolselection}

We adopt the standard PRISMA (Preferred Reporting Items for Systematic Reviews and Meta-Analyses) methodology~\cite{page2021prisma} for conducting and reporting the systematic review and the meta-analysis. \changed{The PRISMA checklist is linked in the supplementary material at the end of the paper. }

The primary objective of our analysis was to examine the application of Artificial Intelligence in the current generation of SLR tools to identify trends and emerging research directions. In order to identify the set of AI-enhanced SLR tools, we first formulated three inclusion criteria and two exclusion criteria. Specifically, the inclusion criteria are the following:
\begin{enumerate}[leftmargin=1.5cm,label={\textbf{IC \arabic*.}},align=left]%
\item The SLR tool must incorporate AI techniques to semi-automate the screening or extraction process, while still maintaining the user's capacity to make the final decision~\cite{tsafnat2014systematic}; 
\item The tool must possess a user interface that facilitates paper screening or information extraction by the user;
\item The tool should not require advanced technical expertise for installation and execution.
\end{enumerate}
The exclusion criteria are:
\begin{enumerate}[leftmargin=1.5cm,label={\textbf{EC \arabic*.}},align=left]
\item The tool is under maintenance;
\item The tool has not been updated in the last 10 years.
\end{enumerate}

\begin{figure}[h]
\centering
\includegraphics[width=\linewidth]{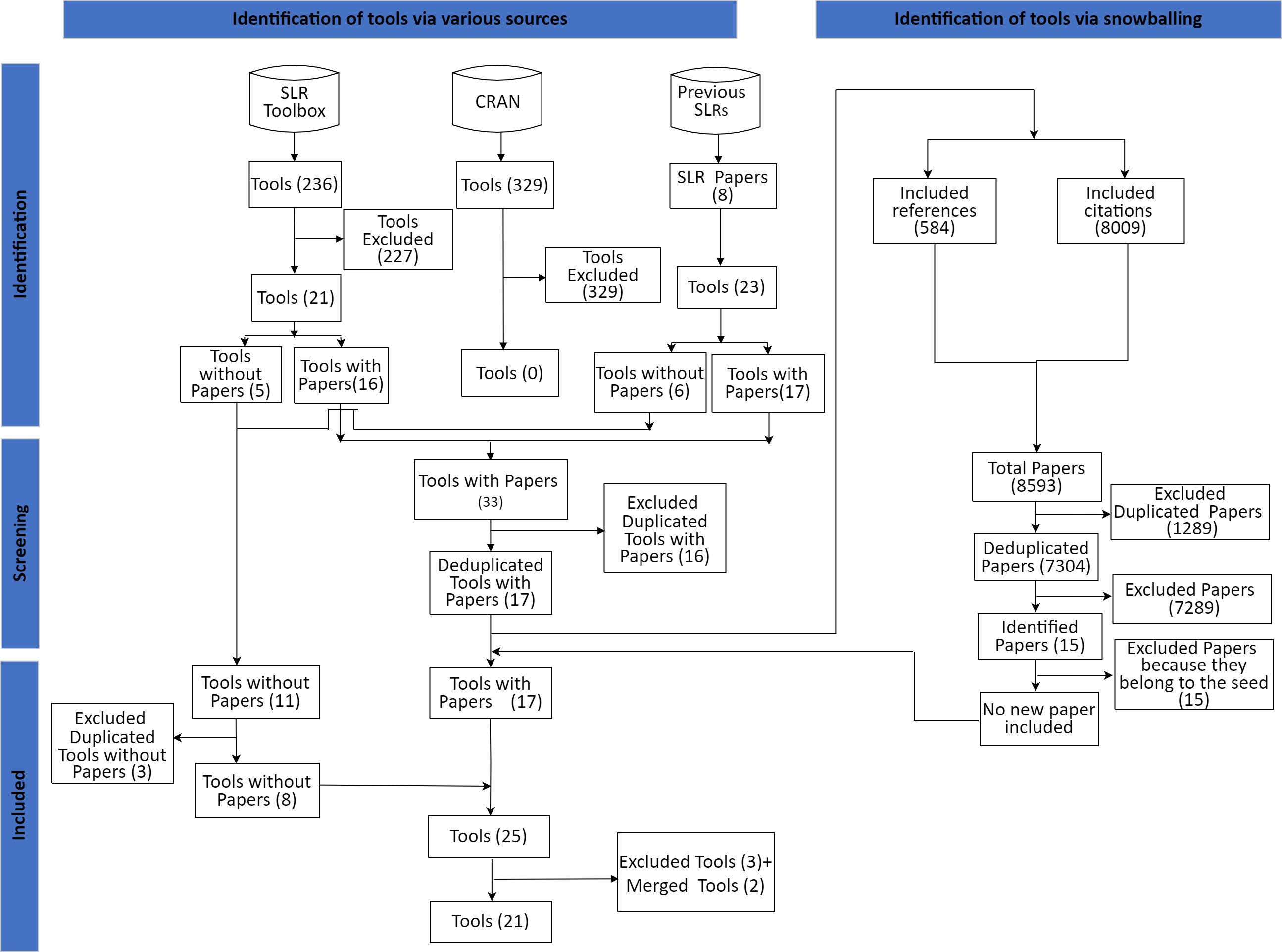}
\caption{PRISMA Diagram of our SLR about AI-enhanced SLR Tools.}
\label{fig:PRISMA}
\end{figure}

The PRISMA diagram, depicted in Figure \ref{fig:PRISMA} illustrates the main phases of the process. 
We utilised three main strategies for identifying the tools. 

\changed{First, we conducted a search of previous survey papers on SLR tools and extracted the tools they analysed. 
This was accomplished using Scopus\footnote{Scopus - \url{https://www.scopus.com/}}, a leading bibliographic database~\cite{publications9010012}. We selected Scopus over other alternatives because it is widely recognised as the preferred source for conducting systematic literature reviews due to its high-quality metadata, reliable citation tracking, and extensive coverage of scientific documents, including journals, conference proceedings, and books \cite{baas2020scopus, visser2021large}.}
Specifically, we used the search string: 
\textit{(``Literature Reviews'' OR ``Systematic Review'') AND (``Tools'' OR ``Automation'' OR ``Semi-Automation'' OR ``Semiautomation'' OR ``Software'')}\footnote{The reader may notice that this query retrieves also documents about ``Systematic Literature Reviews''.}. Since this field lacks standardised vocabulary~\cite{grant2009typology,dieste2009developing}, we aimed to maximise recall by using broad terms, planning to refine the results at a later stage. Additionally, we filtered the results by selecting only `review' as the `Document type' in the Scopus interface.

\changed{
The search yielded 356 review papers. From this set, we identified the surveys focusing on SLR tools. This selection process was conducted in two stages. Initially, the first author, who has eight years of experience in teaching evidence-based medicine and bibliometric analysis, identified a preliminary set of 14 papers based on their titles and abstracts. Subsequently, all four authors collaboratively examined the shortlisted papers by reviewing the full texts. Potentially ambiguous cases were discussed among the authors to achieve consensus. This process yields five survey papers. 
We then applied a snowballing search~\cite{webster2002analyzing} to identify additional surveys. This involved examining the references and citations of the five survey papers. As before, we implemented a two-stage selection process, screening titles and abstracts first, and followed by a full-text analysis of potentially relevant papers. This resulted in the inclusion of three additional papers. 
Overall, this procedure yielded a total of 8 survey papers. An analysis of these surveys led to the identification of 23 tools. 
Among these, 17 were associated with academic papers, whereas 6 were not.}

As a second source, we adopted the Systematic Literature Review Toolbox~\cite{marshall2015systematic}, a repository in which SLR tools are published and updated. \changedtwo{This platform is highly regarded in the field and was adopted as a source in five of the eight previous surveys~\cite{kohl2018online,van2019software,harrison2020software,cowie2022web,robledo2023vista}. Specifically, we utilised the advanced search functionality to retrieve all tools under the ``software'' category.} The query returned 236 tools that were manually analysed. 
Similar to the analysis of the papers, this process was conducted in two phases. Initially, the first author selected a preliminary set of 45 tools based solely on their descriptions in the repository. Next, all authors evaluated these tools by examining their websites, tutorials, and the tools themselves. When necessary, the first author collected additional information through interviews with the developers. As before, any ambiguous cases were discussed among the authors until a consensus was reached.
The analysis identified a total of 21 tools. Of these, 16 were associated with academic papers, while 5 were not.


As a third source, we adopted the Comprehensive R Archive Network (CRAN)\footnote{CRAN repository - \url{https://cran.r-project.org/}}, a well-regarded repository for R packages widely used by the statistical and data science communities. Specifically, we employed the \textit{packagefinder} library~\cite{zuckarelli2019packagefinder} and used the query: \textit{(``systematic literature review'' OR ``systematic review'' OR ``literature review'')}. We chose this library due to its proven effectiveness in retrieving relevant applications in various domains, such as ecology and evolution~\cite{lortie2020checklist}. 
This search yielded 329 tools, which we then evaluated using the same two-stage selection process previously applied for the tools produced by searching the SLR Toolbox. However, none of these tools incorporated AI solutions while also providing a visual interface. Consequently, we discarded all of them.




\changed{
After deduplicating the results from the SLR toolbox and previous surveys, we identified 17 tools linked to research papers and 8 tools without associated papers. To validate and expand our findings, we conducted a snowballing search using the 17 papers linked to the tools as the initial seeds. Our aim was to identify additional papers associated with a relevant tool.}  
\changed{This search was conducted on Semantic Scholar\footnote{Semantic Scholar - \url{https://www.semanticscholar.org/}}, chosen for its extensive coverage in Computer Science, especially for snowballing searches~\cite{hannousse2021searching}. To facilitate this, we developed a custom script to interface with the Semantic Scholar API, enabling the efficient retrieval of references from seed papers and the papers citing them\footnote{Code used for the Snowballing search - \url{https://zenodo.org/records/11154875}}.}
\changed{This process led to the identification of 584 references and 8,009 papers citing the seeds for a total of 8,593 papers. After removing duplicates, 7,304 papers remained. 
The authors analysed these papers using the same two-stage selection process that was employed for survey papers. This analysis yielded 15 of the papers included in the initial seeds as well as the 8 previously identified surveys, but it did not reveal any new tools or papers. The lack of new findings, despite the comprehensive snowballing process, suggests that the tool section identified earlier is exhaustive.}


In conclusion, the process led to a collection of 17 tools with associated papers and 8 without (Covidence, DistillerSR, Nested Knowledge, Pitts.ai, Iris.ai, LaserAI, DRAGON/litstream, Giotto Compliance). We then excluded Giotto Compliance, DRAGON/litstream, and LaserAI from our study due to the lack of available information\footnote{Specifically, there were no associated research papers or comprehensive documentation for these tools, and attempts to contact the developers for further details were unsuccessful.}. We also consolidated RobotReviewer and RobotSearch into a single entry, recognising their shared algorithmic basis. Consequently, the final set of tools considered in our study included 21 distinct tools.


Table \ref{tab:tools_screen_extr} lists the 21 selected tools. The majority of them (17) were primarily designed for the purpose of screening. Two tools, namely Dextr and ExaCT, focused exclusively on extraction. The remaining two tools, Iris.ai and RobotReviewer/RobotSearch, had a dual focus on both screening and extraction.  
In summary, 19 tools can be used for the screening phase and 4 tools for the extraction phase. The majority of the tools (19) are web applications, while only 2, namely SWIFT-Review and ASReview, need to be installed locally. \changed{Furthermore, only 4 tools (Colandr, ASReview, FAST2, and RobotReviewer) release their code under an open license.}




\begin{table}[]
\centering
\footnotesize
\caption{The 21 SLR tools analysed in this survey. OS = Open Scource.\label{tab:tools_screen_extr}}
\begin{tabular}{l|l|l|l|l|c}
\toprule
\textbf{ID} & \textbf{Tool}                & \textbf{Stage SLR} & \textbf{Mode} & \textbf{OS}&\textbf{Reference} \\
\midrule
1 & Abstrackr & Screening & Web & No & ~\cite{wallace2012deploying} \\ 
2 & ASReview & Screening & Desktop & Yes & ~\cite{van2021open} \\ 
3 & Colandr & Screening & Web & Yes & ~\cite{Cheng2018UsingML,cheng2021keep} \\ 
4 & Covidence & Screening & Web & No & - \\ 
5 & DistillerSR & Screening & Web & No & - \\ 
6 & EPPI-Reviewer & Screening & Web & No & ~\cite{thomas2010eppi,2019MachineLF} \\ 
7 & FAST2 & Screening & Web & Yes & ~\cite{yu2019fast2} \\ 
8 & LitSuggest & Screening & Web & No & ~\cite{allot2021litsuggest} \\ 
9 & Nested Knowledge & Screening & Web & No & - \\ 
10 & PICOPortal & Screening & Web & No & ~\cite{agai2020new,minion2021pico} \\ 
11 & Pitts.ai & Screening & Web & No & - \\ 
12 & Rayyan & Screening & Web & No & ~\cite{ouzzani2016rayyan} \\ 
13 & Research Screener & Screening & Web & No & ~\cite{chai2021research} \\ 
14 & RobotAnalyst & Screening & Web & No & ~\cite{przybyla2018prioritising} \\ 
15 & SWIFT-Active Screener & Screening & Web & No & ~\cite{howard2020swift} \\ 
16 & SWIFT-Review & Screening & Desktop & No & ~\cite{howard2016swift} \\ 
17 & SysRev & Screening & Web & No & ~\cite{bozada2021sysrev} \\ 
18 & Dextr & Extraction & Web & No & ~\cite{walker2022evaluation} \\ 
19 & ExaCT & Extraction & Web & No & ~\cite{kiritchenko2010exact} \\ 
20 & Iris.ai & Both & Web & No & - \\ 
21 & RobotReviewer/RobotSearch & Both & Web & Yes & ~\cite{marshall2017automating,marshall2018machine} \\ 
\bottomrule
\end{tabular}
\end{table}

\section{Meta-review of previous surveys}\label{sec:metareview}

This section provides a brief meta-analysis of how the previous systematic literature reviews have described the tools in relation to Artificial Intelligence. We focus on four surveys that analysed AI features~\cite{cowie2022web,de2023artificial,burgard2023reducing,robledo2023vista}. 


The previous survey papers characterised AI according to five main features: 
\begin{enumerate}[leftmargin=1.5cm]
    \item \textbf{Approach} - identifies 
    the method used for performing a specific task. 
    This is the most examined feature, receiving attention from four studies~\cite{cowie2022web,de2023artificial,burgard2023reducing,robledo2023vista}.
    \item \textbf{Text representation} - describes the processes employed to convert text into suitable input for the algorithm (e.g., BoW~\cite{zhang2010understanding}, LDA topics~\cite{blei2003latent}, word embeddings~\cite{wang2020survey}). This feature was analysed by two previous surveys~\cite{de2023artificial,burgard2023reducing}.
    \item \textbf{Human interaction} - specifies how users engage with a tool, detailing the operations and options available to them, as well as the characteristics of the user interface. This is among the least explored features with just one previous study~\cite{de2023artificial}. 
    \item \textbf{Input} - specifies the type of content (full-text or just title and abstract) the tool will need to train its model. Alongside \textit{Human interaction}, this is the least explored feature, with just one previous study~\cite{de2023artificial}.
   \item \textbf{Output} - represents the outcome generated by the trained algorithm, and it has been analysed in three studies~\cite{de2023artificial,burgard2023reducing,robledo2023vista}.\
\end{enumerate}



Table~\ref{tab:aifeatures} summarises the analysis of the four systematic literature reviews and shows how 17 tools have been reviewed according to the five AI features. Only three tools (FASTRED, EPPI-Reviewer, and Abstractr) were actually assessed according to all five AI features.  For ten tools (ASReview, Colandr, Covidence,  DistillerSR, Rayyan, Research Screener, RobotAnalyst, RobotReviewer/RobotSearch, SWIFT-Active Screener, and SWIFT-Review) only three features named \textit{approach}, \textit{text representation}, and \textit{output} have been assessed. The remaining tools were assessed by using only one feature (\textit{approach}).

Upon examining the four survey papers, it is apparent that there is a limited exploration of AI features. De la Torre-López et al.~\cite{de2023artificial} provide the most comprehensive analysis, utilising all five specified features to examine seven tools. In contrast, Burgard et al.~\cite{burgard2023reducing} employed only three features: \textit{text representation}, \textit{approach}, and \textit{output}. Robledo et al.~\cite{robledo2023vista} focused on just two features: \textit{approach} and \textit{output}. Cowies et al.~\cite{cowie2022web} conducted the most restricted analysis, considering only one feature (\textit{approach}). 
Furthermore, the five reported features only offer a narrow perspective on how AI can support SLRs. 

\changed{In summary, the previous systematic reviews offer a relatively limited analysis of the expanding ecosystem of AI-enhanced SLR tools and their characteristics. In the next section, we will address this gap by introducing a comprehensive set of 11 AI features and applying them to evaluate the 21 SLR tools identified in Section~\ref{sec:toolselection}.}


\begin{table}[]
\footnotesize
\caption{
Analysis of SLR tools based on AI features, as conducted by previous surveys~\cite{cowie2022web,de2023artificial,burgard2023reducing,robledo2023vista}. The tools are listed in alphabetical order, with the reviews conducting the analysis cited in the final column. Y = Yes, N=No.\label{tab:aifeatures}}
\begin{tabular}{r|l|c|c|c|c|c|c|l|}
\toprule
\textbf{ID} & \textbf{Tool}                    & \Rot{\textbf{Approach}} & \Rot{\textbf{Text Representation}} & \Rot{\textbf{Human Interaction}}  &  \Rot{\textbf{Input}} & \Rot{\textbf{Output}} & \textbf{Papers}                                  \\
\midrule
1  & Abstrackr               &  \cellcolor[HTML]{C6EFCE}Y     & \cellcolor[HTML]{C6EFCE}Y                 & \cellcolor[HTML]{C6EFCE}Y                             & \cellcolor[HTML]{C6EFCE}Y & \cellcolor[HTML]{C6EFCE}Y  &  ~\cite{cowie2022web,de2023artificial,burgard2023reducing} \\
2  & ASReview                & \cellcolor[HTML]{C6EFCE}Y     & \cellcolor[HTML]{C6EFCE}Y                 & \cellcolor[HTML]{FFC7C4}N             &  \cellcolor[HTML]{FFC7C4}N & \cellcolor[HTML]{C6EFCE}Y  &  ~\cite{burgard2023reducing,robledo2023vista}                 \\
3  & Colandr                 & \cellcolor[HTML]{C6EFCE}Y     & \cellcolor[HTML]{C6EFCE}Y                 & \cellcolor[HTML]{FFC7C4}N             & \cellcolor[HTML]{FFC7C4}N & \cellcolor[HTML]{C6EFCE}Y  &  ~\cite{cowie2022web,burgard2023reducing}                 \\
4  & Covidence               & \cellcolor[HTML]{C6EFCE}Y      & \cellcolor[HTML]{C6EFCE}Y                 & \cellcolor[HTML]{FFC7C4}N             & \cellcolor[HTML]{FFC7C4}N & \cellcolor[HTML]{C6EFCE}Y  &  ~\cite{cowie2022web,burgard2023reducing}                         \\
5  & DistillerSR         & \cellcolor[HTML]{C6EFCE}Y     & \cellcolor[HTML]{C6EFCE}Y                 & \cellcolor[HTML]{FFC7C4}N             & \cellcolor[HTML]{FFC7C4}N & \cellcolor[HTML]{C6EFCE}Y  &  ~\cite{cowie2022web,burgard2023reducing}         \\
6  & EPPI-Reviewer         & \cellcolor[HTML]{C6EFCE}Y     & \cellcolor[HTML]{C6EFCE}Y                 & \cellcolor[HTML]{C6EFCE}Y             & \cellcolor[HTML]{C6EFCE}Y & \cellcolor[HTML]{C6EFCE}Y  &  ~\cite{cowie2022web,de2023artificial,burgard2023reducing} \\
7  & FASTREAD             & \cellcolor[HTML]{C6EFCE}Y     & \cellcolor[HTML]{C6EFCE}Y                 & \cellcolor[HTML]{C6EFCE}Y             & \cellcolor[HTML]{C6EFCE}Y & \cellcolor[HTML]{C6EFCE}Y  &  ~\cite{de2023artificial,burgard2023reducing}                 \\
8  & Giotto   Compliance      & \cellcolor[HTML]{C6EFCE}Y     & \cellcolor[HTML]{FFC7C4}N                 & \cellcolor[HTML]{FFC7C4}N             & \cellcolor[HTML]{FFC7C4}N & \cellcolor[HTML]{FFC7C4}N  &  ~\cite{cowie2022web}                                 \\
9 & Nested   Knowledge    & \cellcolor[HTML]{C6EFCE}Y     & \cellcolor[HTML]{FFC7C4}N                 & \cellcolor[HTML]{FFC7C4}N             & \cellcolor[HTML]{FFC7C4}N & \cellcolor[HTML]{FFC7C4}N  &  ~\cite{cowie2022web}                                 \\
10 & PICOPortal      & \cellcolor[HTML]{C6EFCE}Y     & \cellcolor[HTML]{FFC7C4}N                 & \cellcolor[HTML]{FFC7C4}N             & \cellcolor[HTML]{FFC7C4}N & \cellcolor[HTML]{FFC7C4}N  &  ~\cite{cowie2022web}                         \\
11 & Rayyan   & \cellcolor[HTML]{C6EFCE}Y     & \cellcolor[HTML]{C6EFCE}Y                 & \cellcolor[HTML]{FFC7C4}N             & \cellcolor[HTML]{FFC7C4}N & \cellcolor[HTML]{C6EFCE}Y  &  ~\cite{cowie2022web,burgard2023reducing,robledo2023vista} \\
12 & Research   Screener  & \cellcolor[HTML]{C6EFCE}Y     & \cellcolor[HTML]{C6EFCE}Y                 & \cellcolor[HTML]{FFC7C4}N             & \cellcolor[HTML]{FFC7C4}N & \cellcolor[HTML]{C6EFCE}Y  &  ~\cite{burgard2023reducing}                         \\
13 & RobotAnalyst    & \cellcolor[HTML]{C6EFCE}Y     & \cellcolor[HTML]{C6EFCE}Y                 & \cellcolor[HTML]{FFC7C4}N             & \cellcolor[HTML]{FFC7C4}N & \cellcolor[HTML]{C6EFCE}Y  &  ~\cite{cowie2022web,burgard2023reducing}         \\
14 & RobotReviewer/RobotSearch     & \cellcolor[HTML]{C6EFCE}Y     & \cellcolor[HTML]{C6EFCE}Y                 & \cellcolor[HTML]{FFC7C4}N             & \cellcolor[HTML]{FFC7C4}N & \cellcolor[HTML]{C6EFCE}Y  &  ~\cite{cowie2022web,burgard2023reducing}         \\
15 & SWIFT-Active   Screener & \cellcolor[HTML]{C6EFCE}Y     & \cellcolor[HTML]{C6EFCE}Y                 & \cellcolor[HTML]{FFC7C4}N             & \cellcolor[HTML]{FFC7C4}N & \cellcolor[HTML]{C6EFCE}Y  &  ~\cite{cowie2022web,burgard2023reducing}         \\
16 & SWIFT-Review  & \cellcolor[HTML]{C6EFCE}Y     & \cellcolor[HTML]{C6EFCE}Y                 & \cellcolor[HTML]{FFC7C4}N             & \cellcolor[HTML]{FFC7C4}N & \cellcolor[HTML]{C6EFCE}Y  &  ~\cite{burgard2023reducing,robledo2023vista}         \\
17 & SysRev & \cellcolor[HTML]{C6EFCE}Y     & \cellcolor[HTML]{FFC7C4}N                 & \cellcolor[HTML]{FFC7C4}N             & \cellcolor[HTML]{FFC7C4}N & \cellcolor[HTML]{FFC7C4}N  &  ~\cite{cowie2022web}       \\
\bottomrule
\end{tabular}

\end{table}

\section{Survey of SLR Tools}\label{oursurvey}
We analysed the 21 SLR tools according to 34 features (11 AI-specific and 23 general) by examining the relevant literature (see Table~\ref{tab:tools_screen_extr}), their official websites, and the online tutorials. 
When necessary, we sought additional information by reaching out to the developers through email or online interviews. 

Section~\ref{sec:oursurvey_feature} describes the full set of features, paying particular attention to the new AI features that we first introduced for this survey.   Section~\ref{sec:oursurvey_result} reports the results of the review.  Section~\ref{sec:disc_tool_evolution} discusses the most suitable systems for specific use cases. 
\changed{Finally, Section~\ref{sec:TTV} outlines the threats to validity of our analysis.}


\subsection{Features overview}\label{sec:oursurvey_feature}

\subsubsection{AI features}


To analyse the extent of AI usage within SLR tools, we considered a total of eleven features. 
This evaluation included the five features previously described in Section~\ref{sec:metareview} (approach, text representation, human interaction, input, and output) along with six new features unique to this study. These additional features were identified through a review of the relevant literature~\cite{cowie2022web,de2023artificial,burgard2023reducing,robledo2023vista} and a preliminary analysis of the tools. The six novel features are as follows:

\bigskip

\begin{itemize}[leftmargin=1.5cm]
    \item \textbf{SLR Task} - which categorises the tasks for which the AI approach is used (e.g., paper classification, paper clustering, named-entity recognition); 
    \item \textbf{Minimum requirement} - which refers to the minimum number of relevant and irrelevant papers required to effectively train a classifier tasked with selecting pertinent papers; 
    \item \textbf{Model execution} - which evaluates whether the models operate in real time 
    (synchronously) or later, typically overnight (asynchronously);
    \item \textbf{Research field} - which identifies the research domains in which the tools can be effectively employed; 
    \item \textbf{Pre-screening support} - which specifies  the application of AI techniques to assist users in manually selecting relevant papers, typically by highlighting key terms or grouping similar papers (e.g., topic maps~\cite{howard2020swift} based on LDA~\cite{blei2003latent}, clustering approaches~\cite{przybyla2018prioritising});
    \item \textbf{Post-screening support} - which refers to the application of AI techniques to conduct a final review of the screened papers (e.g., summarisation~\cite{Shah2022TextSU}).
\end{itemize}
\hfill \break
Five of the eleven features (minimum requirement, model execution, human interaction, pre-screening support, and post-screening support) are exclusive to the screening phase and will not be considered when analysing the extraction phase.






\subsubsection{General features}\label{features}

We analysed the non-AI characteristics of SLR tool based on 23 features. We derived these features from previous studies~\cite{marshall2014tools,kohl2018online,van2019software,harrison2020software,cowie2022web} after a process of synthesis and integration. Table~\ref{tab:slr_features_description} shows the description of each feature with its category. To facilitate the systematic analysis, we grouped them into six categories: Functionality (F), Retrieval (R), Discovery (Di), Documentation (Do), Living Review (L), and Economic (E). 

The \textit{functionality} category includes features for auditing and evaluating the technical aspects of the tools. The \textit{retrieval} category covers features related to the acquisition and inclusion of scholarly documents. The \textit{discovery} category consists of features that facilitate the inclusion, exclusion, and management of references during the screening phase. The \textit{documentation} category includes features that support the reporting of the findings. The \textit{living review} category captures the ability of tools to incorporate new relevant documents based on AI techniques. Lastly, the \textit{economic} category reflects the financial considerations associated with the tools. 

\begin{table}[]
\caption{Description of the SLR Features. \label{tab:slr_features_description}}
\footnotesize
\begin{tabular}{p{0.1cm}|p{3.25cm}|p{8.50cm}}
\toprule
\textbf{\#} & \textbf{SLR Feature}                           & \textbf{Description}                                                                                                                    \\
\midrule
1  & Authentication (F)                          & Ability of the tool to authenticate the users involved in the project.                                                       \\
\hline
2  & Multiplatform (F)                     & Ability of the tool to be run on different platforms (e.g., web, desktop).                                                  \\
\hline
3  & Multiple user roles   (F)             & Ability of the tool to allow the user to have different roles (e.g., reviewer, admin) within 
and between projects.          \\
\hline
4  & Multiple user support   (F)           & Ability of the tool to allow multiple users to work on the same project. 
\\
\hline
5  & Project auditing (F)                  & Ability of the tool to track all the changes done in the project.                                                            \\
\hline
6  & Project progress (F)                & Ability of the tool to determine the overall progress of the annotation with respect to the total number of papers to annotate.                                                        \\
\hline
7  & Status of the software   (F)              & The extent to which the tool is actively maintained and has a stable release.                            \\
\hline
8  & Automated full-text retrieval (R)   & Ability of the tool to support full-text retrieval from bibliographic databases.                                             \\
\hline
9  & Automated search (R)                  & Ability of the tool to support literature search through the integration of APIs.                                            \\
\hline
10 & Manual reference importing (R)      & Ability of the tool to allow the user to enter papers manually, typically via a form.                                                          \\
\hline
11 & Manually inserting full-text (R)    & Ability of the tool to allow the user to manually add full-text papers.                                                      \\
\hline
12 & Reference importing   (R)             & Ability of the tool to import papers using a variety of formats (BibTeX, RIS, CSV).                     \\
\hline
13 & Snowballing (R)                         & Ability of the tool to support the automated retrieval of the citations from bibliographic databases (snowballing).            \\
\hline
14 & Deduplication   (Di)                   & Ability of the tool to support the automatic deduplication of the references.                                                       \\
\hline
15 & Discrepancy resolving   (Di)           & Ability of the tool to handle differences of opinion between screeners, e.g., by allowing comments or assigning the problematic papers to a senior screener.  \\
\hline
16 & In-/excluding   references (Di)        & Ability of the tool to allow the user to comment on reference inclusion and exclusion.                                 \\
\hline
17 & Reference labelling   \& comments (Di) & Ability of the tool to allow the user to write additional comments on the references (e.g., 'to double check').                                                \\
\hline
18 & Screening phases   (Di)                & Ability of the tool to allow the user to perform the different stages screening phase.                                         \\
\hline
19 & Exporting results (Do)                 & Ability of the tool to allow exporting the screened references.                              \\
\hline
20 & Flow diagram creation   (Do)           & Ability of the tool to provide the PRISMA diagram of the SLR process.                                                 \\
\hline
21 & Protocol (Do)                          & Ability of the tool to provide the user with pre-defined protocol templates (e.g., the Cochrane guidelines).                                                             \\
\hline
22 & Living/updatable   (L)                & Ability of the tool to update the screened references by automatically including recent and relevant articles.                                  \\
\hline
23 & Free to use (E)                       & It determines whether the tool is available for free or requires payment. \\         
\bottomrule
\end{tabular}
\end{table}

\subsection{Results}\label{sec:oursurvey_result}

In this section, we present the results of our analysis. Section~\ref{sec:oursurvey_AI_Features_Screening} describes the tools for the screening phase through the AI features. Section~\ref{sec:oursurvey_AI_Features_Extraction} presents the tools for the extraction phase also through the AI features. Finally, Section~\ref{sec:oursurvey_SLR_Features} describes the full set of 21 SLR tools according to the general features.

\subsubsection{The Role of AI in the Screening Phase} \label{sec:oursurvey_AI_Features_Screening}


As reported in Table~\ref{tab:tools_screen_extr}, 19 tools use AI for the screening phase. 
In the following, we analyse them according to the  11 AI features. 
To eliminate repetitions, this discussion combines the features \textit{input} and \textit{text representation}  into the category \textit{Input Data and Text Representation}. Moreover, the \textit{output} feature is discussed within the context of the \textit{SLR Task}, as it is contingent upon the specific task requirements. 
The Table in Appendix~\ref{appendixa} reports detailed information on how each of the 19 tools addresses the 11 AI features.

\textbf{Research field.} Twelve tools utilise general AI solutions that are applicable across various research fields. The other seven tools employ specific AI solutions designed to support biomedical studies, typically by identifying Randomised Controlled Trials (RCTs) through the use of dedicated classifiers \cite{noel2021citation}. Notably, EPPI-Reviewer, PICOPortal, and Covidence offer both a general mode and a specialised setting for biomedical studies. Conversely, Pitts.ai, RobotReviewer/RobotSearch, SWIFT-Review, and LitSuggest are exclusively dedicated to the biomedical domain.

\textbf{SLR Task.}  
Fifteen tools utilise artificial intelligence for only one task, most often to classify papers as relevant/irrelevant. The other four tools (Covidence, PICOPortal, and EPPI-Reviewer, Colandr) undertake two AI-related tasks. They all classify papers as relevant/irrelevant, but also execute an additional task, such as identifying a specific type of paper (e.g., economic evaluation, randomised controlled trials, etc.) or categorising papers according to a set of entities defined by the user. 
For the sake of clarity, in our discussion of the subsequent features, we will systematically address the first group (one task) followed by the second group (two tasks). 
In the first group, twelve tools focus on selecting relevant papers given a set of seed papers.  Typically, each paper is assigned an inclusion probability score, usually ranging from 0 to 1.
Of the remaining three, two of them (Pitts.ai and RobotReviewer/RobotSearch) identify RCTs based on a pre-built classification model, while the third (Iris.ai) clusters similar papers to build topic maps that assist users in selecting the relevant papers. 
In the second group, all four systems classify pertinent papers using a set of seed papers as a reference. However, they vary in their secondary AI-driven tasks. Specifically, Covidence and PICOPortal identify RCTs using a predefined classification model. EPPI-Reviewer can identify various types of studies, including RCTs, systematic reviews, economic evaluations, and COVID-19 related studies.
Finally, Colandr, in addition to the standard identification of relevant papers, enables users to define their own set of categories (e.g., ``water management'') and subsequently performs a multi-label classification of articles based on them~\cite{Cheng2018UsingML}.   
It also maps individual sentences to the user-defined categories and provides a confidence score for each classification.

\textbf{AI Approach.}
In the group of tools performing one task, the twelve tools focused exclusively on categorising relevant papers employed various types of machine learning classifiers. 
The most adopted approach is Support Vector Machine (SVM)~\cite{hearst1998support}, which aligns with the findings of prior studies~\cite{schmidt2021data}.
Four of the tools (Abstractr, FAST2, Rayyan, RobotAnalyst) exclusively rely on SVM. Distiller supports both SVM and Naive Bayes. ASReview allows the user to select a vast range of methods, including Logistic Regression, Random Forest, Naive Bayes, SVM, and a Neural Networks classifier. Litsuggest use logistic regression, while SWIFT-Review and  SWIFT-Active Screener use a method based on log-lineal regression. 
Pitts.ai and RobotReviewer/RobotSearch also use an SVM for identifying RCTs~\cite{marshall2018machine}.  
Finally, Iris.ai identifies and groups similar papers based on the similarity of their `fingerprint', a vector representation of the most meaningful words and their synonyms extracted from the abstract~\cite{wuscithon}. 

With regards to the four tools that perform two AI tasks, Covidence, EPPI-Reviewer, and PICOPortal also identify relevant papers by using a SVM classifier. In contrast, Colandr employs a method where it identifies papers by searching for keywords that are related to a set of user-defined search terms~\cite{Cheng2018UsingML}. For instance, it can recognise terms commonly associated with `Artificial Intelligence' and select papers containing these terms.
Covidence also implements a machine learning classifier based on SVM with a fixed threshold for the identification of RCTs following the Cochrane guidelines~\cite{thomas2021machine}. 
EPPI-Reviewer utilises a range of proprietary classifiers trained on various databases to identify papers with distinct characteristics\footnote{EPPI-Reviewer Documentation - \url{https://eppi.ioe.ac.uk/cms/Default.aspx?tabid=3772}}. It uses the Cochrane Randomized Controlled Trial classifier~\cite{thomas2021machine} to identify RCTs.  
It employs a classifier trained with the NHS Economic Evaluation Database (NHS EED)~\cite{craig2007nhs} for identifying economic evaluations and another trained on the Database of Abstracts of Reviews of Effects~\cite{la2004database} to identify systematic reviews in the biomedical field. Finally, it uses a classifier trained on the `Surveillance and disease data on COVID-19'\footnote{COVID surveillance - \url{https://tinyurl.com/229vcpyd}}
for identifying research related to COVID.
PICOPortal employs instead an ensemble of machine learning classifiers, which combines both decision trees and neural networks~\cite{onan2016ensemble}.
Finally, for the identification of the category attributed to the paper by the user, Colandr used a combination of Named Entity Recognition for extracting entities relevant to the categories and a classifier based on logistic regression~\cite{cheng2021keep}. 

\textbf{Input Data and Text representation.}
The AI techniques employed by these tools take as input the title, abstract, or full text of papers. All the tools analysed need only titles and abstracts as input, with the exception of Colandr, which requires the full text of papers. 
The tools generate different representations of the papers to input into the AI models. Specifically, of the 15 tools dedicated to classifying relevant papers, the majority (8 out of 15) use a Bag of Words (BoW) approach~\cite{zhang2010understanding}, while the remainder employ various word embedding techniques~\cite{wang2020survey}. 
Pitts.ai and RobotReviewer/RobotSearch use SciBERT embeddings~\cite{beltagy2019scibert}. Research Screener employs Doc2Vec embeddings~\cite{le2014distributed}. ASReview offers multiple options, including Sentence-BERT~\cite{reimers2019sentence} and Doc2Vec~\cite{le2014distributed}. 
Iris utilises a unique representation called fingerprint~\cite{wuscithon}, which is a vector characterising the most meaningful words and their synonyms extracted from the abstract. 
In the second group, Covidence and EPPI-Reviewer adopt a BoW representation, while PICOPortal uses both BoW and the BioBERT embeddings~\cite{lee2020biobert}. Finally, Colandr uses both word2vec~\cite{mikolov2013distributed} and GloVe~\cite{pennington2014glove} embeddings. Overall, much like other NLP applications, these tools are evolving from traditional text representations like BoW to a range of more modern word and sentence embeddings.



\begin{figure}
     \centering
     \begin{subfigure}[b]{\textwidth}
        \centering
        \includegraphics[width=0.7\linewidth]{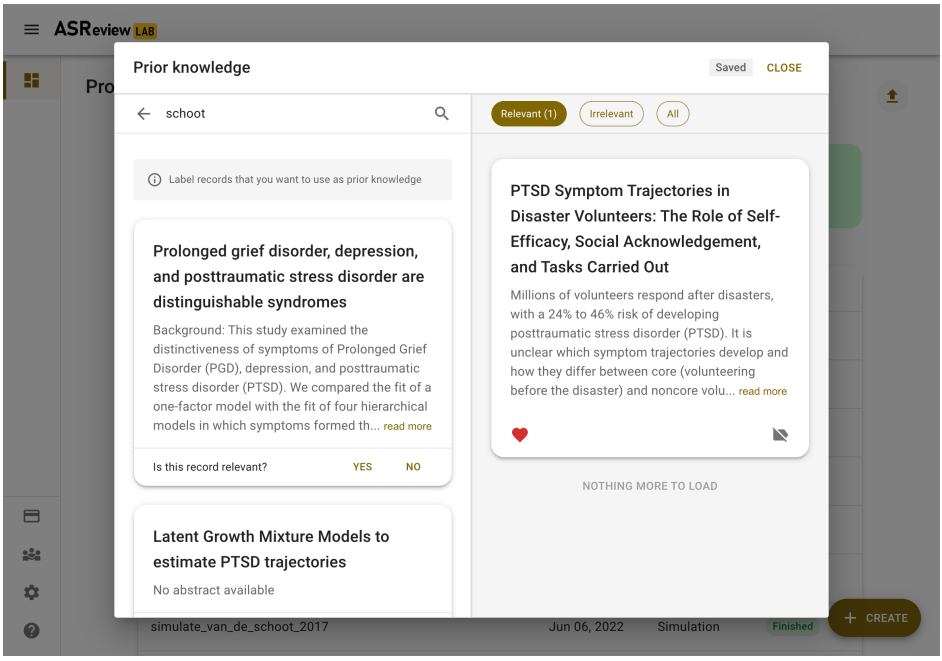}
        \caption{ASReview}
        \label{fig:Interface_Relevant_ASReview}
     \end{subfigure}
     \hfill
     \begin{subfigure}[b]{\textwidth}
        \centering
        \includegraphics[width=0.7\linewidth]{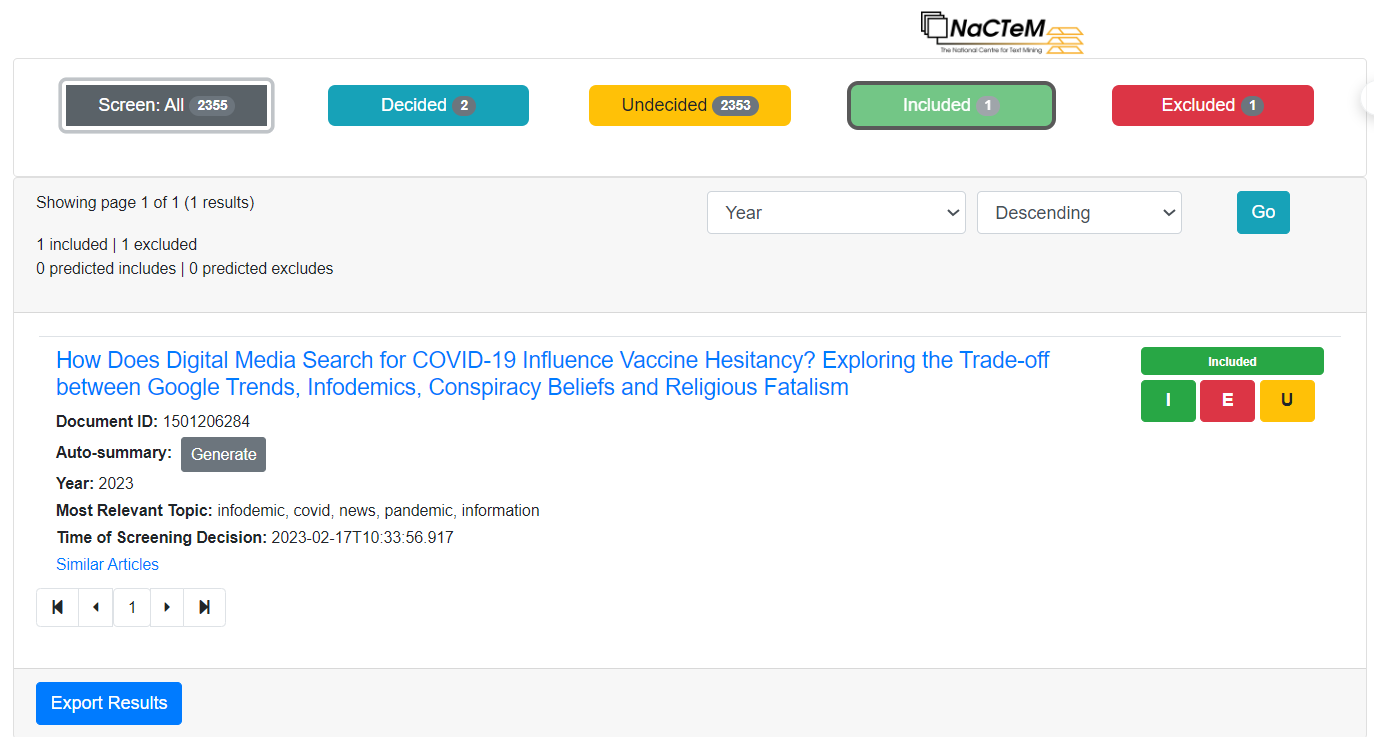}
        \caption{RobotAnalyst}
        \label{fig:Interface_Relevant_RobotAnalyst}
     \end{subfigure}
\caption{Examples of interfaces for paper classification.}
\label{fig:Interface_Relevant}
\end{figure}

\textbf{Human Interaction.} We identified three main types of interfaces. The \textit{first} and most typical one, implemented by 16 tools, regards the classification of paper as relevant. These graphical interfaces typically feature similar templates that allow users to upload and examine papers. Some tools (Rayyan, SWIFT-Active Screener, SysRev, Covidence, and PICOPortal) also offer a menu with additional functionalities like ranking or filtering papers based on specific criteria. 
A few tools (Rayyan, SWIFT-Active Screener, Research Screener, Pitts.ai, SysRev, Covidence, PICOPortal) also enable multiple users to collaboratively perform this task, allowing them to add comments for discussion about problematic papers or to delegate challenging papers to a senior reviewer. 
For illustration, Figure~\ref{fig:Interface_Relevant} (a)  and Figure~\ref{fig:Interface_Relevant} (b) depict the interfaces used by ASReview and RobotAnalyst, respectively, for selecting relevant papers. The ASReview interface enables users to classify papers as either relevant or irrelevant. In contrast, the RobotAnalyst interface provides options for users to categorise papers as included, excluded, or undecided.


\begin{figure}[h]
\centering
\includegraphics[width=\linewidth]{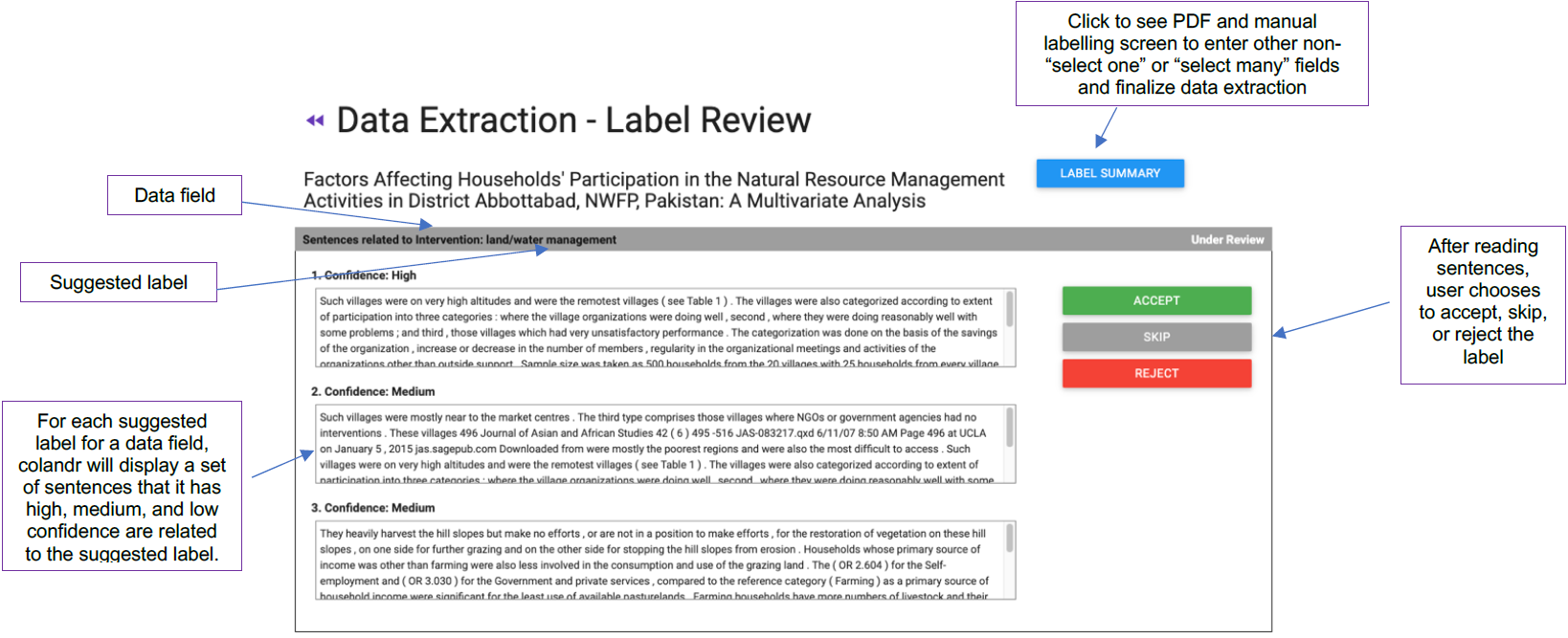}
\caption{Examples of the  tagging process of Colandr. This figure is courtesy of~\cite{colandrgd}.}
\label{fig:Colandr}
\end{figure}

The \textit{second} interface type, offered by Colandr, enables users to define specific categories to assign to the papers. This approach offers greater flexibility compared to the traditional binary classification of relevant or not relevant papers. For instance, in Figure \ref{fig:Colandr} Colandr suggests that for the given paper, the shown sentences are classified with a confidence level of high, medium or low in the category ``land/water management'', previously defined by the user. The user can accept, skip or reject the suggested classification.

\begin{figure}
     \centering
     \begin{subfigure}[b]{\textwidth}
        \centering
        \includegraphics[width=0.8\linewidth]{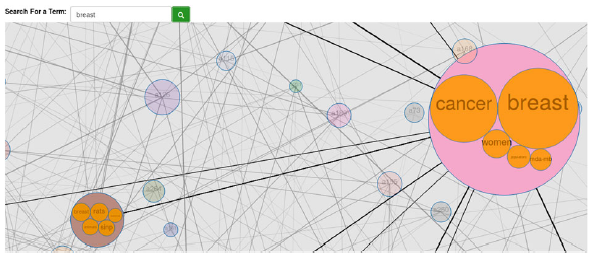}
        \caption{RobotAnalyst: Topic Modelling.}
        \label{fig:RobotAnalyis_TopicModelling}
     \end{subfigure}
     \hfill
     \begin{subfigure}[b]{\textwidth}
        \centering
        \includegraphics[width=0.8\linewidth]{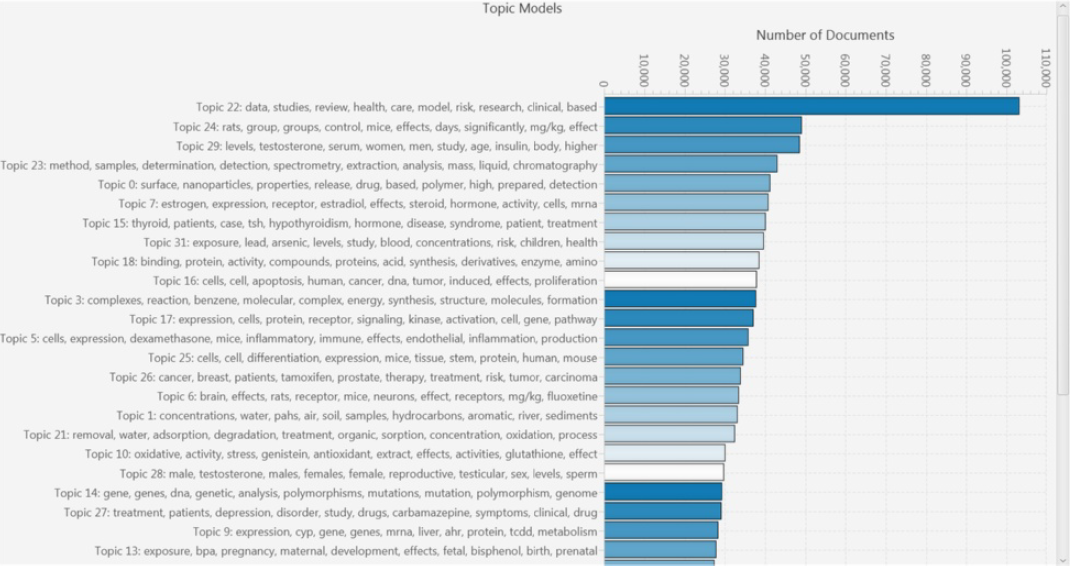}
        \caption{SWIFT-Review: Topic Modelling.}
        \label{fig:SWIFT-Review_TopicModelling}
     \end{subfigure}
     \hfill
     \begin{subfigure}[b]{\textwidth}
        \centering
        \includegraphics[width=0.8\linewidth]{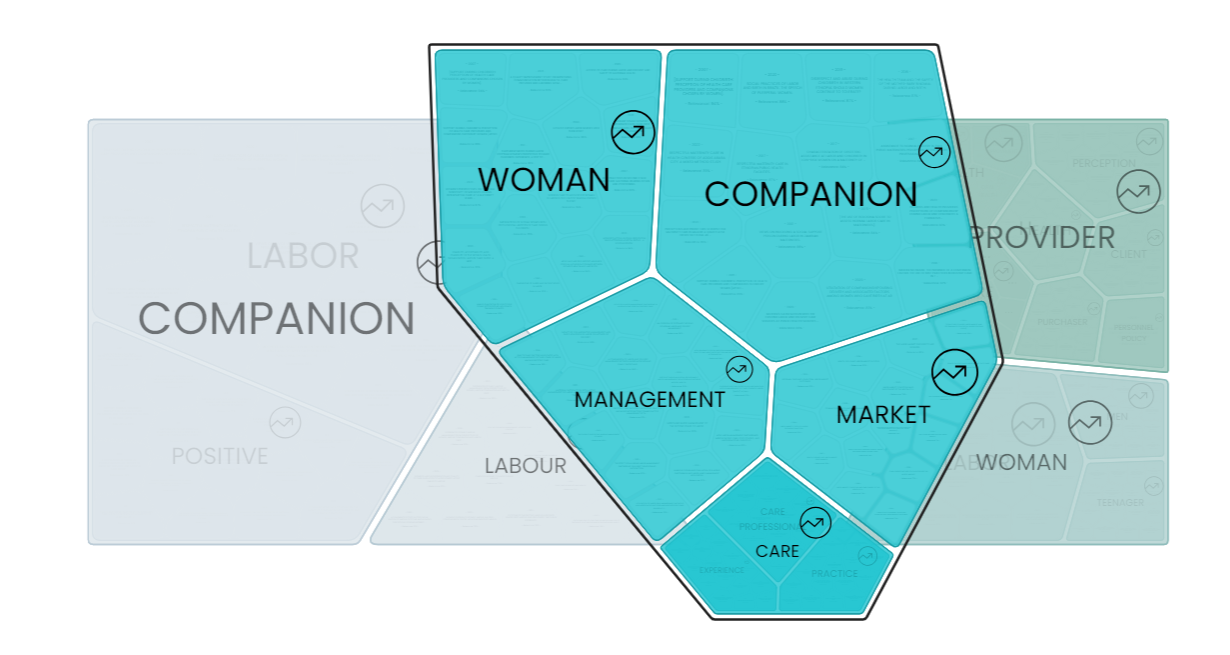}
        \caption{Iris.ai: Specific Topics.}
        \label{fig:IRIS.AI_Screening_Detail}
     \end{subfigure}
\caption{Examples of interactive interfaces for pre-screening.}
\label{fig:SLR_Pre-Screenning_Visualisation}
\end{figure}

The \textit{third} interface type, offered by Iris.ai, is based on a \textit{topic map}, a visualisation technique that clusters papers based on thematic similarities. The user initiates the search process with a brief description of the user's search intent (typically 300 to 500 words), a title, or the abstract of a paper. The system then clusters the papers according to their topics and generates a topic map, such as the ones depicted in Figure \ref{fig:SLR_Pre-Screenning_Visualisation} (c). 
These interactive visualisations enable users to effectively navigate and select papers relevant to their research. 
The user can iteratively repeat the clustering process until they are satisfied that all pertinent papers have been incorporated into the analysis.  
Iris.ai also enables users to further filter the papers according to a variety of facets.


\textbf{Minimum requirements.}  Generally, the accuracy of a classifier improves with an increasing number of annotated papers, but this also increases the time and effort required from researchers. Most methods need between 1-15 relevant papers and typically the same number of irrelevant ones. This is a relatively low number that should allow researchers to quickly annotate the initial set of seed papers.  However, the necessary quantity varies a lot across tools. For instance, ASReview, SWIFT-Active Screener, and SWIFT-Review require just one relevant and one irrelevant paper to begin classification. Covidence and Rayyan require two and five papers, respectively. Other tools require a larger number of papers. For example, Colandr needs 10 seed papers, while SysRev requires 30.



\textbf{Model execution.} Thirteen tools employ a real-time model execution strategy, wherein the training and classification of the model occur immediately after the user selects the relevant and irrelevant paper. Conversely, SysRev and SWIFT-Active Screener adopt a delayed-model-execution approach in which the training and classification steps are conducted at predetermined intervals. Specifically, SysRev executes these operations overnight, whereas SWIFT-Active Screener updates its model after every thirty papers, maintaining a minimum two-minute interval between the most recent and the currently used model.



\textbf{Pre-screening support.}  
Among the 19 tools evaluated, eight implement standard techniques for pre-screening support, such as keyword search, boolean search, and tag search. 
ASReview, Covidence, DistillerSR, and  SWIFT-Active Screener only enable the user to filter the paper by keyword.
Rayyan and EPPI-Reviewer enhance this functionality by highlighting keywords in their visual interface. Additionally, Colandr and Abstrackr offer the feature of colour-coding keywords based on their relevance level.
Rayyan incorporates a boolean search feature, allowing users to combine keywords with operators like \textit{AND}, \textit{OR}, and \textit{NOT}. For example, a boolean search such as \textit{``literature review'' AND ``tools''} will retrieve scholarly documents containing both keywords in their titles or abstracts. Rayyan also provides options to search by author or publication year. EPPI-Reviewer, on the other hand, offers a tag search function, where users can tag papers with specific keywords and then search based on these tags.


\changed{
RobotAnalyst, SWIFT-Review, and Iris.ai also support topic modelling.} The first two use LDA~\cite{blei2003latent}, which probabilistically assigns a topic to a paper based on the most recurrent terms shared by other papers. 
RobotAnalyst presents the topics in a network, as shown in Figure~\ref{fig:SLR_Pre-Screenning_Visualisation} (a) in which each node (circle) represents a topic, and its size is proportional to the frequency of the terms that belong to it. SWIFT-Review uses a simpler approach displaying the topics and their terms in a bar chart, as shown in Figure~\ref{fig:SLR_Pre-Screenning_Visualisation} (b). Iris.ai clusters the papers according to a two-level taxonomy of global topics and specific topics.  
For instance, in Figure~\ref{fig:SLR_Pre-Screenning_Visualisation} (c) we can observe a set of global topics in the background, which include `companion', `labour', `provider', and `woman'. Whereas in the cyan section there are the second-level specific topics, in this case concerning the `labor' global topic, such as `woman', `companion', `market', `management', and `care'. 

RobotAnalyst offers a cluster-based search functionality. This feature employs a spectral clustering algorithm~\cite{ng2001spectral} to group papers. It also incorporates a statistical selection process for identifying the key terms characterising each cluster~\cite{brockmeier2018self}. The resulting clusters are presented to the user, emphasising the most representative terms.

\changed{Finally, Nested Knowledge, PICOPortal, Rayyan, and RobotReviewer/RobotSearch provide PICO identification, which uses distinct colours to highlight the \textit{patient/population}, \textit{intervention}, \textit{comparison}, and \textit{outcome}. 
Rayyan also enhances search capabilities by extracting topics and enriching them with the Medical Subject Headings (MeSH)~\cite{lipscomb2000medical}. Furthermore, it enables users to select biomedical keywords and phrases for inclusion or exclusion. 
}

\begin{figure}
     \centering
     \begin{subfigure}[b]{0.48\textwidth}
        \centering
        \includegraphics[width=\linewidth]{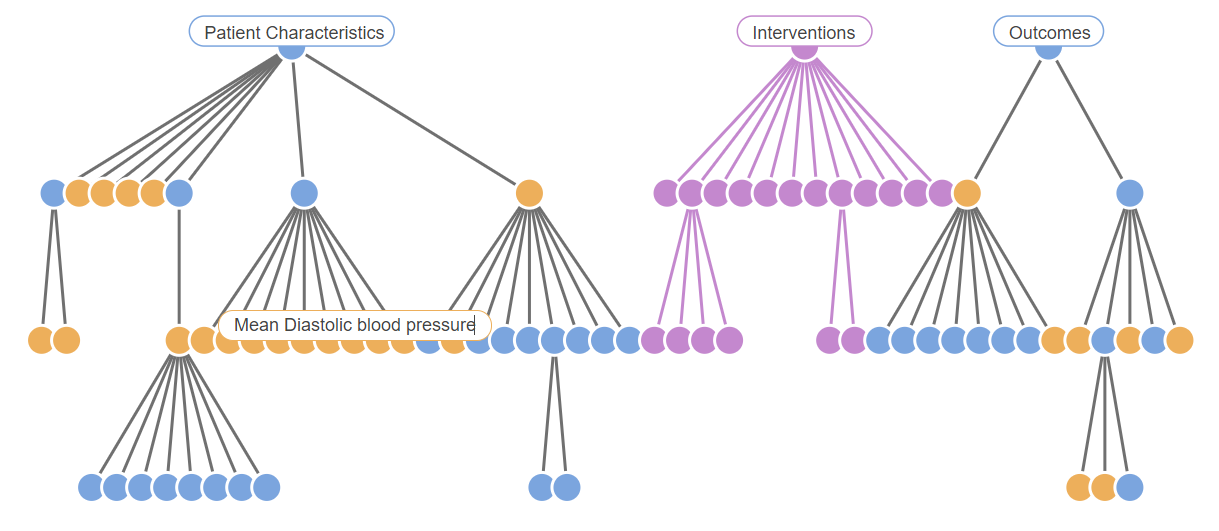}
        \caption{Nested Knowledge: Hierarchical Ontology.}
        \label{fig:NestedKnowgled_Tagging}
     \end{subfigure}
     \hfill
     \begin{subfigure}[b]{0.48\textwidth}
        \centering
        \includegraphics[width=\linewidth]{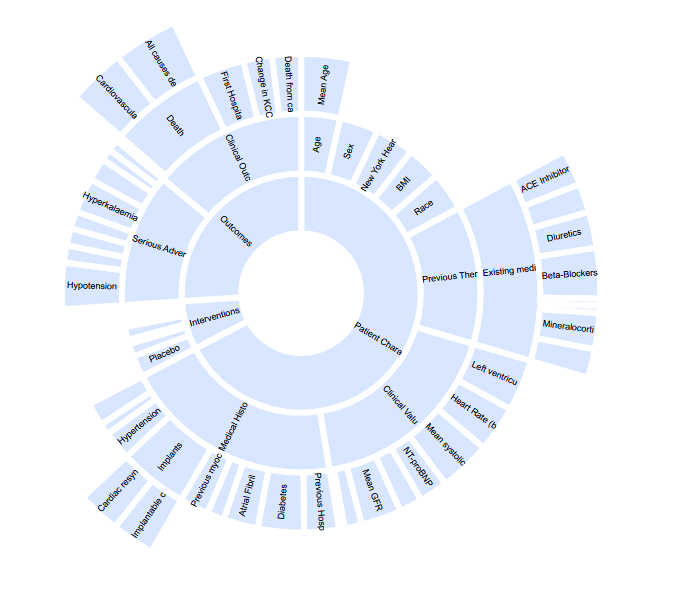}
        \caption{Nested Knowledge: Qualitative Synthesis.}
        \label{fig:NestedKnowgled_Qualitative_Synthesis}
     \end{subfigure}
\caption{Examples of interactive interface for the post-screening.}
\label{fig:SLR_Post-Screenning_Visualisation}
\end{figure}

\textbf{Post-screening support.}  Only two tools offer support for post-screening: Iris.ai and Nested Knowledge. 
Specifically, Iris.ai generates summaries from either a single document, multiple abstracts, or multiple documents. It employs an abstractive summarisation technique~\cite{Shah2022TextSU}, where the summary is formed by generating new sentences that encapsulate the core information of the original text. The system also provides users with the flexibility to adjust the length of the summary, ranging from a brief two-sentence overview to a more comprehensive one-page summary.
Nested Knowledge allows users to create a hierarchy of user-defined tags that can be associated with the documents. For instance, in Figure~\ref{fig:SLR_Post-Screenning_Visualisation} (a), \textit{Mean Diastolic blood pressure} was defined as a sub-tag of \textit{Patient Characteristics}. The user can also visualise the resulting taxonomy as a radial tree chart, as shown in Figure~\ref{fig:SLR_Post-Screenning_Visualisation} (b). 


\subsubsection{The Role of AI in the Extraction Phase}\label{sec:oursurvey_AI_Features_Extraction}

In this section, we describe the four tools that support the extraction phase (Dextr, ExaCT, Iris.ai, and RobotReviewer/RobotSearch) with a focus on the six AI features relevant to the extraction phase. 
We apply the same feature grouping of Section~\ref{sec:oursurvey_AI_Features_Screening}.
The table in Appendix~\ref{appendixb} reports how each of the 4 tools addresses the relevant features.




\textbf{Research Field.} 
RobotReviewer/RobotSearch and ExaCT focus on the medical field, whereas Dextr covers environmental health science. In contrast, Iris.ai can be employed across various research domains.

\textbf{SLR Task.} 
ExaCT, Dextr, and Iris.ai perform Named Entity Recognition (NER)~\cite{nasar2021named} to extract various types of information from the relevant articles. 
Specifically, ExaCT identifies RCT entities based on the CONSORT statement~\cite{moher2001consort}. It returns the top five supporting sentences for each extracted RCT entity, ranked according to relevance. 
Dextr detects data entities used in environmental health experimental animal studies (e.g., species, strain)~\cite{walker2022evaluation}. 
Finally, Iris.ai allows users to customise entity extraction by defining their own set of categories and associating them with a set of exemplary papers.  This is done by filling in a form called Output Data Layout (ODL), which is essentially a spreadsheet detailing all the entities that need to be extracted. 
Finally, RobotReviewer/RobotSearch categorises biomedical articles according to their assessed risk of bias and provides sentences that support these evaluations.



\textbf{AI Approach.} 
The tools perform the NER tasks with a variety of algorithms. 
ExaCT applies a two-step approach~\cite{kiritchenko2010exact}. First, it identifies sentences that are predicted to be similar to those in the pre-trained model, using a SVM classifier. Next, it extracts from these sentences a set of entities via a rule-based approach, relying on the 21 CONSORT categories~\cite{moher2001consort}. Dextr employs a Bidirectional Long Short-Term Memory - Conditional Random Field (BI-LSTM-CRF) neural network architecture~\cite{nowak2019team,walker2022evaluation}. 
Iris.ai does not share specific information about the method used for NER. 
Finally, RobotReviewer/RobotSearch 
employs an ensemble classifier, combining multiple CNN models~\cite{krichen2023convolutional} and soft-margin Support Vector Machines~\cite{boser1992training} 
in order to categorise articles based on their risk of bias assessment (either low or high/unclear) and concurrently extract sentences that substantiate these judgements. The final score for each predicted sentence is the average of the scores obtained from each model.



\textbf{Input Data and Text representation.} 
The majority of the models accept the full-text document as input, except for Dextr, which utilises only titles and abstracts. 
The format requirements vary across these tools. Dextr, Iris.ai, and RobotReviewer/RobotSearch, Iris.ai process papers in PDF format. Dextr also supports input in RIS or EndNote formats. ExaCT encodes papers as HTML.  
The methods for text representation also differ across tools. Dextr encodes text using two pre-trained embeddings: GloVe~\cite{pennington2014glove} (Global Vectors for Words Representations) and ELMo~\cite{Peters2018DeepCW} (Embeddings from Language Models). 
Iris.ai utilises the same fingerprint representation~\cite{wuscithon} discussed in Section~\ref{sec:oursurvey_AI_Features_Screening}. ExaCT uses a simple BoW representation. 
Finally, RobotReviewer/RobotSearch uses BoW for the linear model and an embedding layer for the CNN model.

\subsubsection{General features}\label{sec:oursurvey_SLR_Features}

Table~\ref{tab:slr_features} provides an overview of the proportion of tools covering each of the  23 features.
These features are categorised across the six categories outlined in Section~\ref{features}.
The table in Appendix~\ref{appendixc} provides a more general analysis, detailing how the 21 tools address the 23 features.

The \textit{functionality} category exhibits the highest degree of implementation, with 5 out of 7 features being effectively executed by all the tools. The remaining two features, namely \textit{authentication} and \textit{project auditing},  are implemented by 18 and 9 tools, respectively.  
The other categories present a more heterogeneous scenario. 
Within the \textit{retrieval} category, only \textit{reference importing} is implemented by all tools. Interestingly, no tools provide the ability to automatically retrieve the reference of a paper from bibliographic databases. 
The tools also offer limited support for the feature within the \textit{discovery} category. 
Notably, approximately 50\% of the tools lack basic functionalities such as reference deduplication, options for manual annotation and exclusion of references, and features for labelling and commenting on the references. 
Regarding the \textit{documentation} category, only 4 tools (DistillerSR, Nested Knowledge, Rayyan, and Covidence) provide the PRISMA diagram of the entire SLR process or the protocol templates.
Significantly, LitSuggest stands out as the sole tool providing a \textit{living review}, enabling users to easily update their earlier analyses by automatically adding recent papers that exhibit a high degree of similarity to the previously selected ones. 
In terms of \textit{economic} aspects, the majority of the tools (13 out of 21) are accessible for free.

\begin{table}[]
\caption{Proportion of the 21 tools implementing the 23 generic features.\label{tab:slr_features}}
\begin{tabular}{l|l|l|l}
\toprule
\textbf{Category}              & \textbf{Feature}                   & \textbf{Yes} & \textbf{No} \\
\midrule
\multirow{7}{*}{Functionality} & Multiplatform              & \textbf{21} (100\%)   & 0 (0\%)     \\     
                               & Multiple user   roles              & \textbf{21} (100\%)   & 0 (0\%)     \\
                               & Multiple user   support            & \textbf{21} (100\%)   & 0 (0\%)     \\
                               & Project   auditing                 & 9 (43\%)     & 12 (57\%)   \\
                               & Project   progress                 & \textbf{21} (100\%)   & 0 (0\%)     \\
                               & Authentication                           & 18 (86\%)    & 3 (14\%)    \\
                               & Status of   software               & \textbf{21} (100\%)   & 0 (0\%)     \\ \midrule
\multirow{6}{*}{Retrieval}     & Automated full-text retrieval      & 3 (16\%)     & 16 (84\%)   \\
                               & Automated   search                 & 8 (42\%)     & 11 (58\%)   \\
                               & Snowballing                          & 0 (0\%)      & 19 (100\%)  \\
                               & Manual   reference importing       & 5 (26\%)     & 14 (74\%)   \\
                               & Manually   inserting full-text     & 8 (42\%)     & 11 (58\%)   \\
                               & Reference   importing              & \textbf{21} (100\%)   & 0 (0\%)     \\ \midrule
\multirow{5}{*}{Discovery}     & Deduplication                      & 8 (42\%)     & 11 (58\%)   \\
                               & Discrepancy   resolving            & 12 (57\%)    & 9 (43\%)    \\
                               & In-/excluding   references         & 13 (68\%)    & 6 (32\%)    \\
                               & Reference   labelling \& comments & 10 (53\%)    & 9 (47\%)    \\
                               & Screening   phases                 & 19 (100\%)    & 0 (0\%)    \\ \midrule
\multirow{3}{*}{Documentation} & Exporting results                  & \textbf{21} (100\%)   & 0 (0\%)     \\
                               & Flow diagram   creation            & 4 (21\%)     & 15 (79\%)   \\
                               & Protocol                           & 4 (21\%)     & 15 (79\%)   \\ \midrule
Living   Systematic Review     & Living/updatable                   & 1 (5\%)      & 18 (95\%)  \\ \midrule
Economic                       & Free to use                        & 13 (62\%)    & 8 (38\%)    \\ 
\bottomrule
\end{tabular}
\end{table}

In summary, only eight of the evaluated tools implement at least 70\% of the designated features. Specifically, DistillerSR, Nested Knowledge, Dextr, and ExaCT lead with the highest feature coverage at 82\%. They are followed by PICOPortal and Rayyan, each with 78\%, EPPI-Reviewer with 74\%, and SWIFT-Active Screener with 70\%. Among the remaining 13 tools, eight cover between 50\% and 70\% of the features, while the last five cover between 35\% and 50\%.


\subsection{Outstanding SLR tools}\label{sec:disc_tool_evolution}

Comparing our results with the previous studies in the literature~\cite{marshall2014tools,van2019software,harrison2020software,cowie2022web}, we observed that many SLR tools have undergone significant development and advancements in the last few years. Particularly, the features in the \textit{functionality} category have received more attention and are now considered standard functions. These include capabilities for tracking and auditing projects, multiple user support, and multiple user roles. Additionally, the management of references has seen considerable enhancement.
As discussed in the previous section, the more complete tools in terms of feature coverage include DistillerSR, Nested Knowledge, Dextr, ExaCT,  PICOPortal, and Rayyan. 
However, in practical scenarios, the selection of these tools should be guided by the user's specific needs and use cases. 
\changed{In the following, we provide a brief analysis of some tools that our evaluation has identified as particularly suited to certain scenarios. However, it is important to recognise that there is no single best solution in this complex landscape. Therefore, we encourage researchers to experiment with these tools and determine which ones best meet their requirements.}

In non-biomedical fields, 
ASReviewer stands out for its comprehensive range of methods for selecting relevant articles, including Logistic Regression, Random Forest, Naive Bayes, and Neural Networks classifiers. This makes it a potentially optimal choice for this phase of research. 
Iris.ai and Colandr are also strong contenders that may enable the greatest flexibility since they allow users to respectively cluster documents based on their semantic similarity, and create specific categories for paper classification. Moreover, they offer user-friendly interfaces for analysing the resulting data.
\changed{
Both platforms feature user-friendly interfaces that facilitate the analysis of the resulting data. These features are especially beneficial for exploratory studies aiming to progressively deepen understanding of a domain.}




\changed{
In the biomedical field, Covidence, PICOPortal, EPPI-Reviewer, RobotReviewer/RobotSearch, and Rayyan are all reliable tools.}
Covidence, PICOPortal, and EPPI-Reviewer have also the capability to identify Randomised Controlled Trials (RCTs) using a predefined classification model. 
\changed{
Among these, EPPI-Reviewer offers the most flexibility, since it can be customised to identify a broader range of studies, including systematic reviews, economic evaluations, and research related to COVID-19. 
RobotReviewer/RobotSearch stands out as the only tool that offers automated bias analysis. This feature makes it an ideal choice for researchers who require this specific functionality.
Finally, Rayyan offers a suite of biomedical features, such as PICO highlighting and filtering, the capability to extract study locations, and topic extraction enriched with MeSH terms. It also allows users to define a set of biomedical keywords and phrases for inclusion and exclusion, which is beneficial for identifying specific RCTs.
}

\subsection{Threats to validity}\label{sec:TTV}
This section outlines various threats to the validity of this study. We examined four primary categories of validity threats: internal validity, external validity, construct validity, and conclusion validity~\cite{wohlin2012experimentation}. We considered and mitigated them as follows.

\textbf{Internal Validity.} 
Internal validity in systematic literature reviews concerns to the rigour and correctness of the review's methodology. To ensure the replicability of our review, we meticulously developed a methodologically sound protocol, which incorporated systematic and transparent procedures for the selection of studies and software tools. We also adopted the PRISMA guidelines, known for their robustness and reproducibility (the PRISMA checklist is available in the supplementary material). 
The protocol for this SLR was developed by the first author and reviewed by co-authors to establish a consensus before initiating the review process.  
We identified relevant tools using two prominent software repositories (the Systematic Literature Review Toolbox and the Comprehensive R Archive Network) supplemented by manually searching relevant surveys for additional tools. Additionally, we employed a snowballing search strategy to further extend and validate our results. 
The selection process involved multiple stages to ensure rigorous evaluation and minimise selection bias. Initially, the first author filtered the tools based on the description in the repositories. Next, all authors participated in a more thorough review of the shortlisted tools. In cases where information was unclear or missing, the first author contacted the tool developers directly through email or online interviews. All related publications were thoroughly reviewed to inform the development of the features.
Despite the systematic process, biases could still emerge due to the subjective decisions made by researchers when applying inclusion and exclusion criteria. To mitigate this, we collaboratively reviewed the inclusion or exclusion of the shortlisted tools, thereby reducing the influence of individual biases. 
\changedtwo{Another potential threat to the internal validity arises from the fact that the SLR Toolbox has been offline since March 2024. Although the developers have indicated that it will be operational again soon, there is a possibility that the tool may not be available for future surveys. Nevertheless, we believe that including its results remains valuable, given that this system was utilised in five~\cite{kohl2018online,van2019software,harrison2020software,cowie2022web,robledo2023vista} of the eight previous surveys identified in Section~\ref{sec:toolselection}.}

In conclusion, while the replication of this study by another research team might yield slight variations in the tools and studies included, the robust, systematic methodology employed and the collaborative nature of the review process lend a high degree of internal validity to our findings.



\textbf{External Validity.} 
External validity refers to the degree to which the findings of this systematic literature review are generalisable across various environments and domains. To mitigate threats to external validity, we used multiple sources for selecting the SLR tools. 
Despite these efforts, the selection of search engines and the formulation of search strings might have impacted the completeness of the tool identification. It is possible that some tools were missed because they were not described using the selected keywords or were absent from the targeted repositories and previous surveys. To counteract this limitation, several strategies were implemented. First, search strings were iteratively refined to enhance coverage and ensure a more exhaustive identification of potential tools. Second, a thorough snowballing method was employed. Finally, interviews were conducted with developers of several tools to further ensure the inclusiveness of the tool selection.

Concerning the inclusion and exclusion criteria, we identified two main potential threats to external validity. The first threat stems from the exclusion of tools that do not feature user interfaces. This criterion was set to focus on tools that are readily adoptable by the average researcher. However, earlier studies involving prototypes without interfaces still align with many of our findings. For instance, these studies also conclude that most SLR tools employ relatively outdated AI techniques~\cite{schmidt2023automated,Schmidt2023PreviouslyTE}, as we will discuss more in detail in Section~\ref{sec:ai_screening}. 
The second threat concerns the exclusion of tools that were either under maintenance and unavailable for evaluation or had not been updated in the past ten years. This exclusion criterion might have omitted tools that, despite being inaccessible at the time of the review, could otherwise fulfil the inclusion criteria. These exclusions could potentially restrict the generalisability of our findings. 




\textbf{Construct Validity.} 
Construct validity concerns the extent to which the operational measures used in a study accurately represent the concepts the researchers intend to investigate. In our systematic literature review, a primary concern is whether the 34 features identified to evaluate SLR tools cover all relevant characteristics, particularly concerning the integration of Artificial Intelligence. To address potential gaps identified from previous studies, we developed a set of 11 AI-specific features aimed at capturing aspects previously overlooked. 
Despite these efforts to create a thorough framework for analysis, AI remains a rapidly evolving field, and our feature set might not encapsulate all current and emerging dimensions. To mitigate this issue, the authors collaboratively developed the feature definitions, striving to create a comprehensive representation that incorporates both established dimensions identified in prior surveys and emerging trends noted in recent publications and software developments. Nevertheless, it is acknowledged that some relevant aspects may still be absent from our analysis.



\textbf{Conclusion Validity.} 
Conclusion validity in systematic literature reviews refers to the extent to which the conclusions drawn from the review are supported by the data and are reproducible. In our review, we focused on mitigating threats to conclusion validity by employing a 
systematic process for identifying relevant software tools and extracting pertinent data for analysis. 
To ensure accuracy and consistency in data collection, we developed a data extraction form based on the general and AI-specific features identified during our meta-review and feature analysis. The first author applied this form to a small subset of tools to test its effectiveness. Subsequently, all authors independently used the same form to extract data for the same subset of tools. Comparative analyses of the extracted data revealed a high degree of consistency among authors, thereby validating the data extraction process. Following this validation, the first author continued with the data extraction for the remaining tools. 
Throughout the data analysis and synthesis phases, we engaged in multiple rounds of discussions to refine our categorisation and representation of the features. This collaborative approach aimed to reduce bias and enhance the reliability of our findings.


A persistent threat to conclusion validity in the context of software tool reviews is the dynamic nature of software development~\cite{ampatzoglou2019identifying}. Software tools frequently evolve, acquiring new functionalities that may not be documented in the published literature. To address this, we supplemented our literature review with comprehensive examinations of websites, tutorials, and relevant academic papers. Additionally, we reached out directly to developers to obtain updated or missing information. This proactive approach frequently provided crucial clarifications and additions, which we incorporated into our final review, thereby strengthening the reliability of our conclusions. However, the field of AI is evolving rapidly, particularly in areas such as Generative AI~\cite{brynjolfsson2023generative} and Large Language Models~\cite{min2023recent}. As a result, it is expected that many tools will soon incorporate new AI features. Therefore, while our findings offer a snapshot of the current landscape, they may not fully represent the ongoing advancements.


\section{Research Challenges}\label{sec:challenges}



The current generation of SLR tools can demonstrate significant effectiveness when utilised properly. Nonetheless, these tools still lack crucial abilities, which hampers their widespread adoption among researchers. This section will discuss some of the key research challenges identified from our analysis that the academic community will need to address in future work. 
It is not intended to provide a systematic review like the one in Section~\ref{oursurvey}, but rather to explore some of the most compelling research directions and open challenges, aiming to inspire researchers in this area. 
Section~\ref{sec:ai_screening} analyses the current challenges associated with integrating AI within SLR tools and discusses the potential social, ethical, and legal risks associated with the resulting systems. 
Section~\ref{sec:disc_tool_usage} addresses usability concerns, which represent a major barrier to the adoption of these tools. 
Finally, Section~\ref{sec:evaluation} discusses the challenges in establishing a robust evaluation framework and suggests some best practices.

\subsection{AI for SLR}\label{sec:ai_screening}

As previously discussed, several SLR tools now incorporate AI techniques for supporting in particular the screening and extraction phases. However, current approaches still suffer from several limitations. Consistent with prior research~\cite{kohl2018online,burgard2023reducing,schmidt2021data}, our study reveals that the majority of SLR tools still depend on possibly outdated methodologies. This includes the use of basic classifiers, which are no longer considered state-of-the-art for text and document classification. Likewise, several tools continue to employ BoW methods for text representation, although some of the most recent ones~\cite{walker2022evaluation,van2021open,marshall2018machine} have shifted towards adopting word and sentence embedding techniques, such as  GloVe~\cite{pennington2014glove}, ELMo~\cite{Peters2018DeepCW}, SciBERT~\cite{beltagy2019scibert}, and
Sentence-BERT~\cite{reimers2019sentence}. 
Therefore, the first interesting research direction regards incorporating advanced NLP technologies, particularly the rapidly evolving Large Language Models (LLMs)~\cite{min2023recent}. LLMs represent the state of the art for many NLP tasks and demonstrated remarkable proficiency in classifying and extracting information from documents~\cite{dunn2022structured,xu2023large}. 
However, integrating these models presents several challenges~\cite{ji2023survey}. Firstly, LLMs are trained on general data, resulting in less effective performance in specialised fields and languages with fewer resources. Secondly, LLMs may generate inaccurate or fabricated information, known as ``hallucinations''. Finally, understanding the decision-making process of LLMs is complex, and their outputs can be inconsistent. A possible solution to these issues is the integration of LLMs with different types of knowledge bases that can provide verifiable factual information~\cite{meloni2023integrating}.  This is typically achieved through the Retriever-Augmented Generation (RAG) framework~\cite{lewis2020retrieval}, which allows LLMs to retrieve information from a collection of documents or a knowledge base. 
For example, the recent CORE-GPT~\cite{pride2023core} utilises a vast database of research articles to assist GPT3~\citep{NEURIPS2020_1457c0d6} and GPT4~\cite{openai2023gpt4} in generating accurate answers.
In addition, the extraction phase in particular could be enhanced by also incorporating modern information extraction methods such as event extraction~\cite{li2022survey}, open information extraction~\cite{liu2022open}, and relation prediction~\cite{tagawa2019relation}. 


A second interesting research direction regards interpretability. Indeed, current classification methods for the screening phase typically operate as `black boxes', not giving much additional information on why a certain paper was deemed as relevant. One important research challenge here is to improve this step by including interpretability mechanisms such as fact-checking~\cite{vladika2023scientific} or argument mining~\cite{lawrence2020argument} to provide further insights. Such techniques would provide deeper insights into the screening process, enhancing the reliability and credibility of the tools. In the field of explainable AI~\cite{linardatos2020explainable}, significant research has been conducted to improve our understanding of the processes models use to generate specific outputs. Specifically, in the context of LLMs, various prompting techniques have been developed to enhance the models' ability to explain their reasoning and justify their decisions. These techniques include Chain-of-Thought (CoT)~\cite{wei2023chainofthought}, Tree of Thoughts (ToT)~\cite{yao2023tree,long2023large} and Graph of Thoughts (GoT)~\cite{besta2023graph}.

A third promising research direction involves the use of semantic technologies~\cite{patel2021present}, particularly knowledge graphs, to enhance the characterisation and classification of research papers~\cite{salatino2022cso}. 
Knowledge graphs consist of large networks of entities and relationships that provide machine-readable and understandable information about a specific domain following formal semantics~\cite{peng2023knowledge}. 
They typically organise information according to a domain ontology, which provides a formalised description of entity types and their relationships~\cite{hitzler2021review}. In recent years, we saw the emergence of several knowledge graphs that offer machine-readable, semantically rich, interlinked descriptions of the content of research publications~\cite{jaradeh2019open,salatino2019improving,wijkstra2021living,10.1162/qss_a_00162}.
For instance, the latest iteration of the Computer Science Knowledge Graph (CS-KG)\footnote{Computer Science Knowledge Graph - \url{http://w3id.org/cskg/}} details an impressive array of 24 million methods, tasks, materials, and metrics automatically extracted from approximately 14.5 million scientific articles~\cite{dessi2022}. Similarly, the Open Research Knowledge Graph (ORKG)\footnote{\url{https://www.orkg.org/}} provides a structured framework for describing research articles, facilitating easier discovery and comparison~\cite{jaradeh2019open}. \changed{ORKG currently includes about 25,000 articles and 1,500 comparisons. This survey is also available in ORKG (\url{https://orkg.org/review/R692116}).} In a similar vein, Nanopublications\footnote{\url{https://nanopub.org/}}  allow the representation of scientific facts as knowledge graphs~\cite{groth2010ana}. This method has been recently applied to support ``living literature reviews'', which can be dynamically updated with new findings~\cite{wijkstra2021living}. 
The integration of these knowledge bases offers significant possibilities. It allows for a more detailed and multifaceted analysis of document similarity, and aids in identifying documents related to specific concepts. For instance, it would enable the retrieval of articles that mention particular technologies or that utilise specific materials.





Other SLR phases, such as appraisal and synthesis, received relatively little attention. This gap offers a substantial research opportunity for the application of AI techniques in these areas.
In the appraisal phase, incorporating AI-driven scientific fact-checking tools to evaluate the accuracy of research claims could provide significant benefits~\cite{vladika2023scientific}. For the synthesis phase, the use of summarisation techniques~\cite{altmami2022automatic} and text simplification methods~\cite{sikka2020survey} has the potential to enhance both the efficiency of the analysis and the clarity of the final output. 

Finally, we recommend that the research community participates to scientific events and initiatives in this field, such as ICASR\footnote{International Collaboration for the Automation of Systematic Reviews (\url{https://icasr.github.io/})}~\cite{beller2018making,o2018moving,o2019still,o2020focus}, ALTAR\footnote{Augmented Intelligence for Technology-Assisted Reviews Systems (\url{https://altars2022.dei.unipd.it/})}~\cite{di2022augmented}, and the MSLR Shared Task\footnote{Multidocument Summarisation for Literature Review (\url{https://github.com/allenai/mslr-shared-task})}~\cite{wang2022overview}. These initiatives are focused on discovering the most effective ways in which AI can improve the SLR stages.

\changed{
\subsubsection{AI Impact Assessment}

The importance of evaluating the impact of AI systems has grown significantly, particularly with the recent enactment of the European Commission's Artificial Intelligence Act, which establishes specific requirements and obligations for AI providers. In this context, it is crucial to assess the potential impact of AI-enhanced SLR tools, considering both the relevant literature and the new regulatory framework~\cite{renda2021study,ayling2022putting}.

Stahl et al.~\cite{stahl_systematic_2023} propose an impact assessment model consisting of two main steps: 1) determining whether the AI tool is expected to have a social impact, and 2) identifying the stakeholders who might be affected by the AI system. 
We can apply this model to the SLR tools discussed in this survey.

Regarding social impact, SLR tools aim to support the identification, analysis, and synthesis of findings that are pertinent to specific research questions. 
The information generated by these tools is typically incorporated into research papers and, in some cases, may influence policy development~\cite{birkland2019introduction}. 
The primary concern here is the dissemination of inaccurate scientific information and how such information might be used by the community and policymakers.

Regarding potentially impacted stakeholders, we consider three main groups. The first group consists of authors who use these tools for literature reviews. These individuals face the risk of including incorrect studies and drawing inaccurate conclusions, potentially jeopardising the quality of their work and their careers. To mitigate these risks, it is crucial to use tools that demonstrate high performance and transparency, especially in terms of the datasets used and potential biases. Additionally, these tools should provide mechanisms that allow users to inspect, interpret, and override the tool's choices. 
The second group includes the readers of these literature reviews. They are primarily at risk of being exposed to and subsequently disseminating incorrect or biased information. In addition to the strategies previously mentioned, the scientific community itself plays a crucial role in mitigating this risk by reproducing and correcting earlier results~\cite{munafo2017manifesto}. 
The third group becomes relevant when policy development is involved. In these instances, targeted populations might be affected by policies based on incorrect or biased analyses~\cite{young2005research}. To mitigate this risk, policymakers shall conduct additional analyses to verify the accuracy of the information and use multiple sources. 

In conclusion, while SLR tools carry some inherent risks, these can generally be managed through responsible use and adherence to validation and correction strategies~\cite{myllyaho2021systematic}. A major challenge remains in enhancing the trustworthiness of these tools through robust evaluation mechanisms~\cite{o2019question}. As we will discuss in Section~\ref{sec:evaluation}, the current landscape lacks high-quality evaluation frameworks.

In the context of the recent EU Artificial Intelligence Act\footnote{EU Artificial Intelligence Act - \url{https://www.europarl.europa.eu/RegData/etudes/BRIE/2021/698792/EPRS_BRI(2021)698792_EN.pdf}}, it is important to note that if we classify SLR tools as ``specifically developed and put into service for the sole purpose of scientific research and development'', they would be explicitly exempt from this legislation. Nevertheless, it is still worthwhile to examine how these tools might be categorised under the four risk categories outlined by the AI Act: Unacceptable Risk, High Risk, Limited Risk, and Minimal Risk. After a detailed analysis of the current draft of the legislation, it seems that a typical AI-enhanced SLR tool would most likely be classified as `Limited Risk'. This classification primarily concerns potential issues regarding transparency~\cite{larsson2020transparency}, which may become more pronounced as these tools begin to utilise generative AI~\cite{brynjolfsson2023generative}. 
According to the AI Act, these systems should be ``developed and used in a way that allows appropriate traceability and explainability while making humans aware that they communicate or interact with an AI system as well as duly informing users of the capabilities and limitations of that AI system and affected persons about their rights.''

}

\subsection{Usability}\label{sec:disc_tool_usage}

The current generation of SLR tools remains underutilised~\cite{marshall2018tool}. Most researchers continue to depend on manual methods, often supported by software like Microsoft Excel, or reference management tools~\cite{marshall2015tools} such as Zotero\footnote{Zotero - \url{https://www.zotero.org/}} and Mendeley\footnote{Mendeley - \url{https://www.mendeley.com/}}. Recent studies~\cite{van2019usage}, suggest that this limited usage primarily stems from usability issues, in addition to a few other relevant factors:
\begin{enumerate*}[label=\roman*)]
    \item \textit{steep learning curve}, as researchers may be unfamiliar with the tools' functionalities~\cite{scott2021systematic},
    \item \textit{misalignment with user requirements}, as many of these software deviate from the guidelines set forth by SLR protocols and exhibit limited compatibility with other software systems~\cite{thomas2013diffusion,arno2021views},
    \item \textit{distrust}, as there is uncertainty about the reliability and the mechanisms of these tools~\cite{o2019question,haddaway2020use}, and
    \item  \textit{financial obstacles}, predominantly arising from licensing expenses, along with feature restrictions in trial versions~\cite{dell2021technology}.
\end{enumerate*}
This suggests that usability and accessibility should be prioritised in the design process to encourage wider adoption of these tools~\cite{hassler2014outcomes,hassler2016identification,al2017vision}.

The literature has given limited attention to the usability of SLR tools. To the best of our knowledge, only a few studies focused on this aspect.  For instance, Harrison et al.~\cite{harrison2020software} conducted an experiment where six researchers were tasked with using six different tools in trial projects. 
Findings indicated that two tools also presented in this study, Rayyan and Covidence, were perceived as the most user-friendly. 
Van Altena et al. ~\cite{van2019usage} conducted a survey involving 81 researchers about the usage of SLR tools and found that the primary reasons cited by participants for discontinuing the use of a tool included poor usability (43\%), insufficient functionality (37\%), and incompatibility with their workflow (37\%). 
In the same study, a set of SLR tools was assessed using the System Usability Scale (SUS) questionnaire~\cite{lewis2018system}. 
The tools demonstrated comparable usability, with scores ranging from 66 to 77. These scores correspond to a `C' to `B' grade, indicating satisfactory but not outstanding performance. 

Therefore, a critical challenge in this field lies in the need for more comprehensive research focused on usability. This involves conducting in-depth studies to understand the various aspects of usability, such as effectiveness, efficiency, engagement, error tolerance, and ease of learning~\cite{quesenbery2014five}. The goal is to gather empirical data and user feedback that can provide insights into how users interact with tools, identify common usability issues, and understand the specific needs and preferences of different user groups. Based on these findings, it is essential to develop robust, evidence-based usability guidelines~\cite{schall2017usability}. These guidelines should offer clear and actionable recommendations for designing user-friendly interfaces and functionalities in future tools. 

\changed{
\subsection{Evaluation of SLR tools}\label{sec:evaluation}
A robust evaluation framework is essential for comparing SLR tools and supporting their continuous improvement~\cite{national2019reproducibility}. 
In the following subsections, we will first discuss the shortcomings of existing evaluation methods and then propose a set of best practices as an initial step towards developing a high-quality evaluation framework.


\subsubsection{Lack of Standard Evaluation Frameworks}

The assessment of SLR tools presents a significant challenge due to the absence of standard evaluation frameworks and established benchmarks. }
Existing literature includes various evaluations of SLR tools that focus on individual phases of the SLR process~\cite{burgard2023reducing,liu2018comparative,yu2018finding}. 
However, these evaluations are not directly comparable due to variations in datasets and evaluation methodologies. 
Moreover, most SLR tools are tested using small, custom datasets, which may not provide a realistic representation of their performance in typical usage scenarios~\cite{burgard2023reducing}.
Additionally, leading commercial providers of SLR tools typically do not make evaluation data available, which complicates comparisons with both existing competitors and new prototypes developed by the research community.

Another concern is related to the performance metrics. Indeed, canonical metrics like precision, recall, and F1-score may not suffice to assess these tools.
For instance, for the screening phase, it is critical to minimise the costs of screening while preserving a high recall. For this reason, it was suggested to adopt the F2 score~\cite{9552024} instead of the F1 score. The F2 score is computed as the weighted harmonic mean of precision and recall. In contrast with the F1 score, which assigns equal importance to precision and recall, the F2 score places greater emphasis on recall compared to precision. 
The Work Saved over Sampling (WSS)~\cite{van2021open}  is another metric that proved to be quite effective in assessing the screening phase~\cite{burgard2023reducing}.
However, Kusa et al.~\cite{kusa2022evaluation} point out that this measure depends on the number of documents and the proportion of relevant documents in a dataset, making it difficult to compare the performance of different screening tasks performed over different systematic reviews. To address this, they introduced the Normalised Work Saved over Sampling (nWSS) metric~\cite{kusa2023analysis}, which facilitates the comparison of paper screening performance across various datasets. 

Another limitation arises from the restricted range of dimensions assessed during the evaluation of SLR tools. \textit{Performance} is only one of several aspects that should be considered. \textit{Usability}, as discussed in Section~\ref{sec:disc_tool_usage}, is another crucial factor. 
Trustworthiness is also a vital dimension~\cite{o2019question}. Although trustworthiness is partially reliant on performance, it also involves reliability, transparency, and ethical integrity, all of which can influence researchers' willingness to use these tools. Indeed, while automation might boost the efficiency of the review process, it also carries the risk of introducing errors.  
These errors could lead to the omission of pertinent studies or the inclusion of inappropriate ones, which could substantially alter the research results. 
To address these issues, O'Connor et al.~\cite{o2019question} propose two main strategies to enhance trust in SLR tools. The first strategy is to undertake detailed studies comparing the precision of automated tools with conventional review methods. The second strategy involves encouraging reputable teams or funding agencies to support the use of these tools.
Wang et al.~\cite{wang2023investigating} recommend that creators of AI-driven tools should investigate different affordances to enhance user trust. Specifically, they identify three essential design elements: i) clear communication about AI capabilities to set appropriate user expectations; ii) availability of user settings to adjust and tailor preferences related to AI-generated recommendations; and iii) inclusion of indicators that explain the mechanisms of the underlying models, helping users evaluate the AI's suggestions. 
Bernard et al. \cite{bernard2023systematic} further expand on this by advocating for the assessment of fairness, accountability, transparency, and ethics (FATE) aspects. They also explore definitions, approaches, and evaluation methodologies aimed at developing trustworthy information retrieval systems. 
A promising avenue for future research is to further explore the application of these concepts to the emerging generation of SLR tools and, more in general, to AI tools designed to support research activities.

In the realm of AI-enhanced SLR tools, \textit{transparency} is one of the greatest challenges in building user trust. 
This is primarily because most contemporary AI models operate as black boxes, making their internal processes difficult to comprehend~\cite{castelvecchi2016can}. Additionally, as shown in Table 2, only 4 out of the 21 tools analysed operate under open licenses, which exacerbates the lack of transparency.
To mitigate this issue, existing research suggests several approaches. These include making the AI model, its training data, and the corresponding code openly accessible for examination by users and experts~\cite{national2018open,Abbasip1609}. Furthermore, it is recommended that developers offer thorough evaluations of any potential biases and perform ablation studies to determine common error types~\cite{ntoutsi2020bias}. It is also advised to integrate explainable AI methods~\cite{linardatos2020explainable}.





\subsubsection{Towards an Evaluation Framework: Best Practices}

In this section, we aim to present some best practices to establish a robust evaluation framework for SLR tools, informed by our surveys and subsequent analysis. While developing a comprehensive evaluation framework is beyond the scope of this paper, we aim to contribute to ongoing discussions by proposing an initial theoretical framework.

We propose a set of best practices centred around three critical aspects of SLR systems: \textit{performance}, \textit{usability}, and \textit{transparency}. First, we suggest developing replicable methodologies to assess the performance of various algorithmic components designed to address the tasks outlined in the previous section. Second, we recommend conducting a comprehensive and unbiased assessment of usability. Finally, we emphasise the importance of improving the trustworthiness of these tools by disclosing essential information about their capabilities and limitations, sharing knowledge bases and models, and adopting explainable AI solutions.

We do not claim that this set of principles is exhaustive; rather, it represents an initial effort to introduce a few principles that could make evaluations in this domain more reproducible and transparent. The principles we outline precede specific implementation decisions and are not tied to any particular technology, standard, or method. 
It is also important to recognise that these principles are not novel but reflect established guidelines used by various communities facing similar challenges~\cite{liu2018comparative,o2019question,wang2023investigating,bernard2023systematic,national2018open,Abbasip1609,linardatos2020explainable}. However, as noted previously, the community that develops SLR tools does not consistently adhere to these practices, leading to a lack of comparability among these systems. 


The proposed best practices are outlined in the following.

\bigskip
\textbf{Performance.}
All models and algorithms employed by an SLR tool for specific tasks should undergo formal evaluation. These evaluations must adhere to established benchmarks and best practices recognised by the relevant scientific community. 
We thus recommend the following practices.
\begin{enumerate}
  \item \textit{Detailed Documentation}: 
  Provide a comprehensive description of all the algorithms employed for the different functions within the system.
  \item \textit{Standardised Evaluation}: Evaluate these algorithms against standard metrics and benchmarks that are widely accepted within the scientific community relevant to the tasks being performed.
  \item \textit{Benchmark Disclosure}: Publicly release the benchmarks used for evaluating these methods to facilitate comparison with alternative approaches.
  \item \textit{Benchmark Adoption}: Whenever possible, opt to reuse established benchmarks, especially those that are recognised and have previously been used for evaluating SLR tools.
  \item \textit{Code Availability}: Ensure that the code for both the algorithms and the evaluation process are persistently available on an online repository to promote accessibility and reproducibility.
\end{enumerate}

\bigskip
\textbf{Usability.}
The evaluation of usability should be comprehensive, replicable, and conducted in environments that closely resemble the diverse settings in which the system will operate, involving various types of potential users. To ensure a thorough assessment, we recommend the following practices.
\begin{enumerate}
  \item \textit{Representative User Participation}:  Conduct detailed user studies with participants who accurately represent the system's target user base.
  \item \textit{Diverse Usability Factors}: The user studies should comprehensively evaluate various usability aspects discussed in the literature such as effectiveness, efficiency, engagement, error tolerance, and ease of learning.
  \item \textit{Standard Questionnaires}: To facilitate comparisons with other systems, the evaluation should also employ established usability questionnaires such as the System Usability Scale (SUS)~\cite{brooke1996sus}, the User Experience Questionnaire (UEQ)~\cite{laugwitz2008construction}, or the Usability Metric for User Experience (UMUX)~\cite{kraig2010}. 
  \item \textit{Accessibility}: 
  Evaluate usability for individuals with diverse disabilities, including visual, auditory, physical, speech, cognitive, language, learning, and neurological. Adopting the Web Content Accessibility Guidelines (WCAG) 2.1, developed by the World Wide Web Consortium (W3C), is recommended to guide this process.
  \item \textit{Availability of Materials}: Publish all materials related to the usability evaluation in a third-party repository to ensure reproducibility.
\end{enumerate}

\bigskip
\textbf{Trasparency.}
In line with the AI Act and the necessity of enhancing user trust~\cite{o2019question}, transparency is essential for AI-driven SLR tools. It is important to incorporate transparency also in the evaluation process, ensuring traceability and explainability, and clearly defining the tools’ capabilities and limitations. Although proprietary systems might emphasise confidentiality to preserve a competitive advantage, it is crucial to balance commercial interests with the broader imperative for accountability and trust in AI technologies. We recommend the following practices to enhance transparency.

\begin{enumerate}
  \item \textit{Availability of Training Data}:  Since the training dataset influences the model's behaviour and can perpetuate biases, ensuring its availability is essential.
  \item \textit{Availability of Knowledge Bases}:  Many systems utilise various knowledge bases, such as taxonomies and vocabularies of research areas, to enhance performance. These resources should be made accessible for user inspection.
  \item \textit{Availability of Models}: Trained models should be made available to facilitate further analysis of their performance and potential biases.
  \item \textit{Explainability}: The tool should, wherever possible, provide clear explanations for its decisions, aligning with the principles of explainable AI.
  \item \textit{Comprehensive Documentation}: All functionalities of the software should be documented clearly and in user-friendly language.
  \item \textit{Clarify the Limitations}: Developers should clearly communicate the limitations of the software, indicating where the tool is expected to perform well and where it may not meet expectations.
\end{enumerate}

\bigskip

We aim for these best practices to serve as an initial step in establishing a comprehensive evaluation framework. 
We hope that this effort will be expanded through dedicated theoretical and empirical research, promoting wider implementation of recognised best practices within this field.


\section{Emerging AI Tools for Literature Review}\label{sec:other_ai_tools}
\changedtwo{Since 2023, a new generation of AI tools designed to assist researchers has emerged.} This development is largely influenced by the advancements in Large Language Models~\cite{sanderson2023ai}. 
Several leading bibliographic search engines are currently introducing LLM  technology. For instance, Scopus and Dimensions\footnote{Dimensions- \url{https://www.dimensions.ai/}} are working on their own chatbot engine and are planning to release them throughout 2024~\cite{van2023chatgpt,aguilera2024scopus}. Similarly, CORE\footnote{CORE - \url{https://core.ac.uk/}}, a search engine providing access to 280 million papers, has recently presented the prototype CORE-GPT, an enhanced version that can answer natural language queries by extracting information from these documents~\cite{pride2023core}. 
These LLM-based tools do not directly support specific SLR phases as the applications that we reviewed in Section~\ref{oursurvey}. Nevertheless, their functionalities can aid researchers in conducting literature reviews and are expected to be integrated into future SLR tools. 
Therefore, when discussing the advancement of AI-driven SLR tools and identifying research challenges in this domain, it is essential to consider these tools and their features. 
A comprehensive analysis of emerging LLM-based tools designed to assist with literature reviews and scientific writing would require an extensive survey. This section aims to present an initial exploratory study that provides insights into how this new generation of LLM-based tools is being used to assist research and what functionalities could potentially be integrated into SLR tools. Since this is an exploratory study rather than a systematic review, we adopted a straightforward search strategy focusing on tools available as online services. Therefore, we excluded tools that are solely described in academic papers and not available for practical use. 

We used TopAI Tools\footnote{TopAI Tools- \url{https://topai.tools/}}, a renowned search engine indexing more than 11K AI systems and searched for the following relevant terms: ``literature review'', ``systematic review'', ``scientific research'', ``search engine'', and ``writing assistant''. 
This search returned 164 tools, which were processed using the same two-stage selection process described in Section~\ref{sec:toolselection}. We first screened the tools by using their short descriptions and then all authors performed a thorough examination of 18 candidate tools. 
This process yielded 11 tools in this domain. Table \ref{tab:LiteratureReviewTools} reports an overview of these tools.

The eleven systems that we identified typically employ LLMs (mostly via the OpenAI API\footnote{OpenAI API - \url{https://openai.com/blog/openai-api}}) often enhanced with a RAG framework~\cite{lewis2020retrieval} to integrate knowledge from scientific and technical documents. 
As discussed in Section~\ref{sec:ai_screening}, the RAG framework enhances LLMs by enabling them to retrieve relevant information from a knowledge base or a collection of documents. This information is then incorporated into the context of the LLMs, allowing them to rely on verifiable sources and thereby reducing inaccuracies and hallucinations~\cite{ji2023survey}. 

We classified the 11 tools into two categories: search engines and writing assistants. Search engines enable users to enter a query using natural language and provide a list of related research papers and their summaries. Their main contribution is the ability to use natural language rather than keywords for searching research papers. On the other hand, writing assistants accept a description of a document, such as ``Survey paper about knowledge graphs'', and generate pertinent text that can be then iteratively refined by a researcher. 
Seven tools were categorised as search engines and three as writing assistants. Textero.ai was the only identified tool fitting into both categories.


\begin{table}[]
\caption{Literature Review Tools based on LLMs.\label{tab:LiteratureReviewTools}}

\begin{tabular}{p{0.2cm}| p{2.3cm}|p{1.2cm}| p{2.3cm}|p{4.2cm}}
\toprule
\textbf{ID} & \textbf{Tool}                & \textbf{Mode} & \textbf{Type} & \textbf{Website} \\
\midrule
1           & Scite~\cite{nicholson2021scite,Rife2021sciteTN,brody2021scite}                    & Web                 & Search Engine    & \href{https://scite.ai/}{https://scite.ai/}                \\
2           & Elicit~\cite{kung2023elicit}                   & Web                  & Search Engine    & \href{https://elicit.com/}{https://elicit.com/}               \\
3           & Consensus                  & Web              & Search Engine    & \href{https://consensus.app/}{https://consensus.app/}                      \\
4           & EvidenceHunt                  & Web               & Search Engine    & \href{https://evidencehunt.com/}{https://evidencehunt.com/}              \\
5           & MirrorThink                  & Web               & Search Engine    & \href{https://mirrorthink.ai/}{https://mirrorthink.ai/}           \\
6           & Perplexity                  & Web/App                 & Search Engine    & \href{https://www.perplexity.ai/}{https://www.perplexity.ai/}                      \\
7           & Scispace                  & Web                   & Search Engine    & \href{https://typeset.io/}{https://typeset.io/}                       \\
8           & Jenni.ai                  & Web/App                 & Writing Assistant    & \href{https://jenni.ai/}{https://jenni.ai/}                      \\ 
9           & ResearchBuddies                  & Web                & Writing Assistant    & \href{https://researchbuddy.app/}{https://researchbuddy.app/}                       \\ 
10         & Silatus                  & Web             & Writing Assistant    & \href{https://silatus.com/}{https://silatus.com/}                       \\ 
11          & Textero.ai                  & Web                  &Both     & \href{https://textero.ai/}{https://textero.ai/}                       \\

\bottomrule

\end{tabular}
\smallskip\footnotesize
    
\end{table} 

\subsection{Search Engine Tools}

The tools in this category allow users to formulate a natural language query and generate a list of relevant research papers sourced from online repositories. Generally, these tools also provide concise summaries of the most prominent papers. Beyond the natural language query functionality, some tools incorporate additional search features. For instance, EvidenceHunt allows users to locate papers using keywords, medical specialisations, or filters specific to PubMed searches. Similarly, Scite offers the capability to conduct keyword searches in titles and abstracts, and uniquely, to search for specific terms within `citation statements'~\cite{nicholson2021scite}, i.e., segments of text that include a citation~\cite{ding2014content}.
Additionally, Scispace and Elicit allow users to automatically extract information from papers based on predefined categories. 
For instance, a user can request the extraction of all references to `technologies' within a text. However, the quality of the extracted results can vary significantly.

The bibliographic databases employed by these tools differ. Elicit, Consensus, and Perplexity utilise Semantic Scholar\footnote{Semantic Scholar - \url{https://www.semanticscholar.org/}}. EvidenceHunt relies on PubMed\footnote{PubMed - \url{https://pubmed.ncbi.nlm.nih.gov/}}. Scite sources its content from Semantic Scholar and a broader array of publishers, such as Wiley, Sage, Europe PMC, Thieme, and Cambridge University Press.  The bibliographic databases used by Scispace, Textero.ai, and MirrorThink are not documented. 
Scite and Consensus process full-text papers, while Elicit and EvidenceHunt only use titles and abstracts.

The majority of the tools (6 out of 8) are versatile and applicable across different research fields. EvidenceHunt is specifically tailored for use in biomedicine, while Elicit is designed to cater to both biomedicine and social sciences.

The specific details of the implementation for many of these tools remain undisclosed, as they are proprietary commercial products. However, it appears that a majority of them employ the OpenAI API, utilising various prompting strategies and often integrating a RAG framework~\cite{lewis2020retrieval} to incorporate text from pertinent articles. Notably, two of the tools explicitly state their models: Elicit and Perplexity; both of which leverage OpenAI's GPT technology.





\subsection{Writing Assistant Tools}

These tools enable the user to describe the document they want to generate and then iteratively refine it.  
Jenni.ai is a highly interactive tool that enables collaborative editing between the user and the AI. Initially, the user provides a step-by-step description of the desired text. Subsequently, the system generates a template for the document and progressively incorporates new sections. These sections can be edited by the user in real-time, facilitating a dynamic and iterative writing process. 
Textero.ai operates similarly. Users are required to input the title and description of the text they wish to create. They can then request the tool to gather pertinent references for integration and select a citation style, such as MLA or APA. The generated text can be further refined by the user either manually or through various AI functions designed to enhance or summarise sections of the text. Additionally, a panel on the right side provides convenient access to the list of cited references, with each paper accompanied by a brief summary. For this reason, we categorised this tool also as a search engine. 
Silatus can operate in four distinct modes: \textit{question answering}, which generates a specific answer; \textit{research report}, producing a comprehensive explanation of a research topic; \textit{blog post}, creating content suitable for blogs; and \textit{social media post}, tailored for social media platforms. In each mode, the user is prompted to provide a concise initial prompt to initiate text generation. Optionally, the user can instruct Silatus to retrieve and integrate pertinent references into the generated text.

As before, most systems do not disclose their technologies, yet they appear to incorporate different versions of the OpenAI API, augmented with specific prompting techniques. Silatus employs GPT-4, while Jenni.ai uses a combination of GPT-3.5 and its proprietary AI technologies. It remains unclear whether any of them have fine-tuned their models for writing-related tasks. 

The quality of the text produced by these systems varies significantly, even when using very informative prompts. Presently, these tools may be more beneficial for master's students who are required to write brief essays rather than for researchers. Nonetheless, as the technology continues to evolve, it is anticipated that a new generation of tools will emerge, offering substantial assistance in academic writing. 
These AI systems could potentially automate several complex tasks, such as generating comprehensive literature reviews~\cite{hope2023computational}, recommending citations~\cite{ali2020deep,buscaldi2024citation}, and identifying new scientific hypotheses~\cite{sybrandt2020agatha,borrego2022completing}. 



\section{Conclusion}\label{conclusion}

In this survey, we performed an extensive analysis of SLR tools, with a particular focus on the integration of AI technologies in the screening and extraction phases. Our study includes a detailed evaluation of 21 tools, examining them across 11 AI-specific features and 23 general features. The analysis extended to 11 additional applications that leverage LLMs to aid researchers in retrieving research papers and supporting the writing process. 
Throughout the survey, we critically discussed the strengths and weaknesses of existing solutions, identifying which tools are most suitable for specific use cases. We also explored the main research challenges and the emerging opportunities that AI technologies present in this field. 

Our findings paint an exciting picture of the current state of SLR tools. We observed that the existing generation of tools, when used effectively, can be highly powerful. However, they often fall short in terms of usability and user-friendliness, limiting their adoption within the broader research community. 
Concurrently, a new generation of tools based on LLMs is rapidly developing. While promising, these tools are still in their infancy and face challenges, such as the well-documented issue of hallucinations in LLMs. 
This highlights the need for the research community to focus on knowledge injection 
and RAG 
strategies to ensure the generation of robust and verifiable information.

The challenges identified in our survey represent a vibrant and evolving area of interest for researchers. It is anticipated that in the next five years, we may see the emergence of a novel generation of AI-enabled research assistants based on LLMs. These AI-enabled research assistants could support researchers by performing a variety of crucial tasks such as generating comprehensive literature reviews, 
identifying new scientific hypotheses, 
and fostering crucial innovation in research practices. 
The research community bears the crucial task of steering the growth of AI, minimising bias, and upholding strict ethical standards. 
With the AI revolution impacting many fields, it is essential to remember that human critical thinking and creativity 
are still vital and remain a core responsibility of the researchers.

\backmatter

\changed{
\bmhead{Supplementary material}
The full versions of the tables describing the 21 SLR tools according to the 34 features are accessible online on both GitHub (\url{https://angelosalatino.github.io/ai-slr/}) and the Open Research Knowledge Graph (\url{https://doi.org/10.48366/R692116}). 
The PRISMA Checklist is available at \url{https://angelosalatino.github.io/ai-slr/\#PRISMA_Checklist}. 
A high-resolution version of all figures is also available at \url{https://angelosalatino.github.io/ai-slr/\#paper_figures}.  
The code used for conducting the snowballing search is archived on Zenodo and can be accessed at \url{https://zenodo.org/records/11154875}.
}




\bmhead{Acknowledgments}
We would like to express our gratitude to the developers of the following tools for providing additional information via email or in personal interviews: Covidence, DistillerSR, Nested Knowledge, Pitts.ai, PICOPortal, Rayyan, SWIFT-Active Screener, SWIFT-Reviewer, and SysRev.










\begin{appendices}

\section{Systematic Literature Review Tools analysed through AI and Generic Features}\label{appendix}

In this appendix, we report three tables that describe the 21 systematic literature review tools examined according to both generic and AI-based features. In Appendix~\ref{appendixa} and Appendix~\ref{appendixb}, we present the analysis of the AI features for the screening and the extraction phases, respectively. In Appendix~\ref{appendixc}, we report the analysis of the tools according to the generic features.
Due to space constraints, only a summarised version of these tables is included here. 
\changed{The full version is available online on both GitHub (\url{https://angelosalatino.github.io/ai-slr/}) and the Open Research Knowledge Graph (\url{https://doi.org/10.48366/R692116}).}

\pagestyle{empty}
\newgeometry{margin=1cm}
\begin{landscape}

\newpage
\subsection{Screening Phase of Systematic Literature Review Tools analysed through AI Features}\label{appendixa}

{\scriptsize
\rowcolors{3}{white}{orange!20}
\begin{longtable}{L{3cm}|L{2cm}|L{4cm}|L{4cm}|L{4cm}|L{4cm}}
\textbf{Tool} & \textbf{Research Field} & \textbf{SLR Task} & \textbf{Text Representation} & \textbf{Input} & \textbf{Minimum Requirement}  \\ \hline 
\endhead
Abstrackr                 & Any            & Classification of relevant   papers.                                                                                                                                                                                          & Bag of words.                                                                                                                                                                                      & Title \& Abstract                                                                                    & -                                                                                                                                                 \\
ASReview                  & Any            & Classification of relevant   papers.                                                                                                                                                                                          & Bag of words.        Embeddings: SentenceBERT, doc2vec.                                                                                                                                            & Title \& Abstract                                                                                    & Relevant papers: 1.        Irrelevant papers: 1.                                                                                                  \\
Colandr                   & Any            & \textbf{Task 1}: Classification of   relevant papers.        \textbf{Task 2}: Identification of the category attributed to the paper by the user.                                                                                               & \textbf{Task 1}: Embeddings:   Word2vec.        \textbf{Task 2}: Embeddings: Glove                                                                                                                                   & \textbf{Task 1}: Title \&   Abstract        \textbf{Task 2}: Full content                                              & \textbf{Task 1}: 10 relevant papers and 10   irrelevant papers.        \textbf{Task 2}: Minimum 50 papers.                                                          \\
Covidence                 & Any            & \textbf{Task 1}: Classification of   relevant papers.        \textbf{Task 2}: Identification of biomedical studies (RCTs).                                                                                                                      & Bag of words for both tasks:   ngrams.                                                                                                                                                             & \textbf{Task 1}: Title \&   Abstract        \textbf{Task 2}: Title \& Abstract                                         & \textbf{Task 1}: 2 relevant papers and 2   irrelevant papers.        \textbf{Task 2}: Not Applicable.                                                               \\
DistillerSR               & Any            & Classification of relevant   papers.                                                                                                                                                                                          & Bag of words.                                                                                                                                                                                      & Title \& Abstract                                                                                    & Relevant papers: 10.        Irrelevant papers: 40.                                                                                                \\
EPPI-Reviewer             & Any            & \textbf{Task 1}: Classification of   relevant papers.        \textbf{Task 2}: Identification of biomedical studies (RCTs, Systematic Reviews,   Economic Evaluations, COVID-19 categories, long COVID).                                         & \textbf{Task 1}: Bag of words   (ngrams).        \textbf{Task 2}: The Cochrane RCT classifer uses bag of words. For the other   approaches the information is not available.                                         & \textbf{Task 1}: Title \&   Abstract        \textbf{Task 2}: Title \& Abstract                                         & \textbf{Task 1}: 5 relevant papers. Number   of irrelevant papers not available.        \textbf{Task 2}: Not Applicable                                             \\
FAST2                     & Any            & Classification of relevant   papers.                                                                                                                                                                                          & Bag of words.                                                                                                                                                                                      & Title \& Abstract                                                                                    & -                                                                                                                                                 \\
Iris.ai                   & Any            & Clustering of Abstracts                                                                                                                                                                                                       & Embeddings.                                                                                                                                                                                        & Title \& Abstract                                                                                    & Not Applicable                                                                                                                                    \\
LitSuggest                & Biomedicine    & Classification of relevant   papers.                                                                                                                                                                                          & Bag of words.                                                                                                                                                                                      & Title \& Abstract                                                                                    & -                                                                                                                                                 \\
Nested   Knowledge        & Any            & Classification of relevant   papers.                                                                                                                                                                                          & -                                                                                                                                                                                                  & Title \& Abstract                                                                                    & -                                                                                                                                                 \\
PICOPortal                & Any            & \textbf{Task 1}: Classification of   relevant papers.        \textbf{Task 2}: Identification of biomedical studies (RCTs).                                                                                                                      & Embeddings for \textbf{Task 2}:   BioBERT.        No information regardin \textbf{Task 1}.                                                                                                                           & \textbf{Task 1}: Title \&   Abstract        \textbf{Task 2}: Title \& Abstract                                         & -                                                                                                                                                 \\
pitts.ai                  & Biomedicine    & Identification of biomedical   studies (RCTs).                                                                                                                                                                                & Embeddings: SciBERT                                                                                                                                                                                & Title \& Abstract                                                                                    & Not Applicable                                                                                                                                    \\
Rayyan                    & Any            & Classification of relevant   papers.                                                                                                                                                                                          & Bag of words: ngrams                                                                                                                                                                               & Title \& Abstract                                                                                    & Relevant papers: 5.        Irrelevant papers: 5.                                                                                                  \\
Research   Screener       & Any            & Classification of relevant   papers.                                                                                                                                                                                          & Embeddings: paragraph embedding                                                                                                                                                                    & Title \& Abstract                                                                                    & Relevant papers: 1.        Irrelevant papers: Information not available.                                                                          \\
RobotAnalyst              & Any            & Classification of relevant   papers.                                                                                                                                                                                          & Bag of words.                                                                                                                                                                                      & Title \& Abstract                                                                                    & -                                                                                                                                                 \\
RobotReviewer/RobotSearch & Biomedicine    & Identification of biomedical   studies (RCTs).                                                                                                                                                                                & Embeddings: SciBERT                                                                                                                                                                                & Title \& Abstract                                                                                    & Relevant papers: NA.        Irrelevant papers: NA.                                                                                                \\
SWIFT-Active   Screener   & Any            & Classification of relevant   papers.                                                                                                                                                                                          & Bag of words.                                                                                                                                                                                      & Title \& Abstract                                                                                    & Relevant papers: 1.        Irrelevant papers:1.                                                                                                   \\
SWIFT-Review              & Biomedicine    & Classification of relevant   papers.                                                                                                                                                                                          & Bag of words.                                                                                                                                                                                      & Title \& Abstract                                                                                    & Relevant papers: 1.        Irrelevant papers:1.                                                                                                   \\
SysRev.com                & Any            & Classification of relevant   papers.                                                                                                                                                                                          & -                                                                                                                                                                                                  & Title \& Abstract                                                                                    & Relevant papers: 30.        Irrelevant papers: 30.                                                                                                  \\ \hline  
\end{longtable}
}

\newpage
\subsection{Extraction Phase of Systematic Literature Review Tools analysed through AI Features}\label{appendixb}
{\scriptsize
\rowcolors{3}{white}{orange!20}
\begin{longtable}{L{2cm}|L{2cm}|L{3cm}|L{6cm}|L{3cm}|L{2cm}|L{4cm}}
\textbf{Tool} & \textbf{Research Field} & \textbf{SLR Task} & \textbf{Approach} & \textbf{Text Representation}  & \textbf{Input}   & \textbf{Output}   \\      \hline 
\endhead
RobotReviewer / RobotSearch & Biomedical                   & Identifies risks of bias: how   reliable are the results?                                                                                         & ML classifier, combining a   lineal model and a Convolutional Neural Network (CNN) model.       These models are trained on a dataset containing manually annotated   sentences stating the level of bias.                            & Bag of word: ngrams.      Embeddings: embedding layer from CNN Model. & Full-text paper.    & Risk of bias classification (as   Low, High, Unclear)        \\
ExaCT                     & Biomedical                   & NER of Randomised Controlled   Trials                                                                                                             & \textbf{Task 1}: ML classifier based on   SVM to identify sentences regarding a control trial.      \textbf{Task 2}: Rule base detection to identify the 21 CONSORT categories.                                                                         & Bag of words: ngrams.                                                                                           & Full-text paper.    & Possible RCT entities        \\   
Dextr                     & Environmental Health Science & \textbf{Task 1}: NER of animal   studies.      \textbf{Task 2}: Entity linking of animal studies.                          & \textbf{Task 1}: ML Classifier   implementing a neural network model based on bidirectional LSTM with a   Conditional Random Field (BI-LSTM-CRF) architecture.      \textbf{Task 2}: Linking according to a customised ontology                         & \textbf{Task 1}: Embeddings: GloVe,   ELMo.      \textbf{Task 2}: Not Applicable.       & Title and Abstracts & \textbf{Task 1}: Possible animal   entities.      \textbf{Task 2}: Relationships of animal models and exposures vs experimentas ot   endpoints vs experiments. \\
Iris.ai                   & Any                          & \textbf{Task 1}: NER of entities selected   by the user.      \textbf{Task 2}: Entity linking of the identified entities. & \textbf{Task 1}: ML classifier.   Algorithim is unknown.      \textbf{Task 2}: Uses a knowledge graph to represent the relations of within the   entities on the paper or between the entities of the table. The technical   implementation is unknown. & \textbf{Task 1}: Embeddings: word   embedding.      \textbf{Task 2}: Not Applicable.    & Full-text paper.    & \textbf{Task 1}: Possible entities based   on a confidence interval.      \textbf{Task 2}: Additional semantics on the extracted entities.   \\       \hline 
\end{longtable}
}

\newpage
\subsection{Systematic Literature Review Tools analysed based on General Features}\label{appendixc}


{\scriptsize
\rowcolors{3}{white}{orange!20}
\begin{longtable}{L{2.4cm}|l|l|l|l|l|c|c|c|c|c|c|c|c|c|c|c}
\Rott{\textbf{Tool}} &\Rott{\textbf{Multiple user roles}} &\Rott{\textbf{Multiple user support}} &\Rott{\textbf{Project auditing}} &\Rott{\textbf{Automated full-text retrieval}} &\Rott{\textbf{Automated search}} &\Rott{\textbf{Snowballing}} &\Rott{\textbf{Manual reference importing}} &\Rott{\textbf{Manually inserting full-text}} &\Rott{\textbf{Deduplication}} &\Rott{\textbf{Discrepancy resolving}} &\Rott{\textbf{In-/excluding references}} &\Rott{\textbf{Reference labelling \&   comments}} &\Rott{\textbf{Flow diagram creation}} &\Rott{\textbf{Protocol}} &\Rott{\textbf{Living/updatable}} &\Rott{\textbf{Free to use}} \\ \hline 
\endhead
Abstrackr                  & Sing.               & 2                     & Yes              & No                            & None                                         & No          & Yes                        & No                           & No            & Yes                   & No                       & Yes                               & No                    & No       & No               & Yes         \\
Colandr                    & Sing.               & 2                     & No               & No                            & None                                         & No          & No                         & No                           & No            & Yes                   & Yes                      & Yes                               & No                    & Yes      & No               & Yes         \\
DistillerSR                & Mult.               & \textgreater{}1       & Yes              & Yes                           & PubMed                                       & No          & Yes                        & Yes                          & Yes           & Yes                   & Yes                      & Yes                               & Yes                   & No       & No               & No          \\
EPPI-Reviewer              & Mult.               & \textgreater{}1       & Yes              & No                            & PubMed                                       & No          & Yes                        & Yes                          & Yes           & Yes                   & Yes                      & Yes                               & No                    & No       & No               & No          \\
LitSuggest                 & Sing.               & No                    & No               & No                            & PubMed                                       & No          & No                         & No                           & No            & No                    & No                       & No                                & No                    & No       & Yes              & Yes         \\
Nested Knowledge           & Mult.               & \textgreater{}1       & Yes              & Yes                           & PubMed; Europe PMC; DOAJ; ClinicalTrials.gov & No          & No                         & Yes                          & Yes           & Yes                   & Yes                      & Yes                               & Yes                   & Yes      & No               & No          \\
Rayyan                     & Mult.               & \textgreater{}1       & Yes              & No                            & None                                         & No          & Yes                        & Yes                          & Yes           & Yes                   & Yes                      & Yes                               & Yes                   & No       & No               & Yes         \\
RobotAnalyst               & Sing.               & No                    & No               & No                            & PubMed                                       & No          & Yes                        & No                           & No            & No                    & Yes                      & No                                & No                    & No       & No               & Yes         \\
SWIFT-Active Screener      & Mult.               & \textgreater{}1       & Yes              & No                            & None                                         & No          & No                         & Yes                          & Yes           & Yes                   & Yes                      & Yes                               & No                    & Yes      & No               & No          \\
SWIFT-Review               & Sing.               & No                    & No               & No                            & None                                         & No          & No                         & No                           & No            & No                    & No                       & No                                & No                    & No       & No               & Yes         \\
FAST2                      & Sing.               & No                    & No               & No                            & None                                         & No          & No                         & No                           & No            & No                    & No                       & No                                & No                    & No       & No               & Yes         \\
ASReview                   & Sing.               & \textgreater{}1       & No               & No                            & None                                         & No          & No                         & No                           & No            & No                    & Yes                      & No                                & No                    & No       & No               & Yes         \\
Research Screener          & Mult.               & \textgreater{}1       & No               & No                            & None                                         & No          & No                         & No                           & Yes           & Yes                   & Yes                      & No                                & No                    & No       & No               & Yes         \\
pitts.ai                   & Mult.               & \textgreater{}1       & No               & No                            & PubMed                                       & No          & No                         & No                           & No            & Yes                   & Yes                      & No                                & No                    & No       & No               & No          \\
SysRev.com                 & Mult.               & \textgreater{}1       & Yes              & No                            & PubMed                                       & No          & No                         & Yes                          & No            & Yes                   & Yes                      & Yes                               & No                    & No       & No               & No          \\
Covidence                  & Mult.               & \textgreater{}1       & No               & No                            & None                                         & No          & No                         & Yes                          & Yes           & Yes                   & Yes                      & Yes                               & Yes                   & No       & No               & No          \\
RobotReviewer /RobotSearch & Sing.               & No                    & No               & No                            & None                                         & No          & No                         & No                           & No            & No                    & No                       & No                                & No                    & No       & No               & Yes         \\
Iris.ai                    & Sing.               & No                    & Yes              & No                            & CORE; PubMed; US Patent Office; CORDIS       & No          & No                         & No                           & No            & No                    & No                       & No                                & No                    & No       & No               & No          \\
PICO Portal                & Mult.               & \textgreater{}1       & Yes              & Yes                           & None                                         & No          & No                         & Yes                          & Yes           & Yes                   & Yes                      & Yes                               & No                    & Yes      & No               & Yes         \\
Dextr                      & Sing.               & No                    & No               & NA                            & None                                         & NA          & NA                         & NA                           & NA            & No                    & NA                       & NA                                & NA                    & NA       & NA               & Yes         \\
ExaCT                      & Sing.               & No                    & No               & NA                            & None                                         & NA          & NA                         & NA                           & NA            & No                    & NA                       & NA                                & NA                    & NA       & NA               & Yes          \\ \hline

\end{longtable}

}

\end{landscape}

\end{appendices}
\restoregeometry

\newpage

\bibliography{sn-bibliography}

\end{document}